\PassOptionsToPackage{table}{xcolor}
\documentclass[letterpaper]{article} 
\usepackage{aaai2026}  
\usepackage{times}  
\usepackage{helvet}  
\usepackage{courier}  
\usepackage[hyphens]{url}  
\usepackage{graphicx} 
\usepackage{natbib}  
\usepackage{caption} 
\usepackage{algorithm}
\usepackage{algorithmic}
\usepackage{amsmath}
\usepackage{amssymb} 
\usepackage{booktabs}
\usepackage{multirow}
\usepackage{diagbox}
\usepackage{times}
\usepackage{graphicx}
\usepackage{helvet}
\usepackage{courier}
\usepackage{amssymb}
\usepackage{natbib} 
\usepackage{indentfirst}
\usepackage{tcolorbox}
\usepackage{courier} 
\tcbuselibrary{listings}
\usepackage[table]{xcolor}
\usepackage{colortbl}
\usepackage{float}   
\usepackage{enumitem}
\usepackage{url}
\usepackage{colortbl}

\usepackage{newfloat}

\DeclareCaptionStyle{ruled}{labelfont=normalfont,labelsep=colon,strut=off} 
\floatstyle{ruled}
\newfloat{listing}{tb}{lst}{}
\floatname{listing}{Listing}
\title{T3Time: Tri-Modal Time Series Forecasting via Adaptive Multi-Head Alignment and Residual Fusion}
\author{
    Abdul Monaf Chowdhury\textsuperscript{\rm 1}, 
    Rabeya Akter\textsuperscript{\rm 1}, 
    Safaeid Hossain Arib\textsuperscript{\rm 1}
}

\affiliations{
    \textsuperscript{\rm 1}Robotics \& Mechatronics Engineering, University of Dhaka\\
    \{monafabdul15, rabeyaakter231023, safaeid48\}@gmail.com

}

\usepackage{bibentry}

\begin{document}

\maketitle

\begin{abstract}
Multivariate time series forecasting (MTSF) seeks to model temporal dynamics among variables to predict future trends. Transformer-based models and large language models (LLMs) have shown promise due to their ability to capture long-range dependencies and patterns. However, current methods often rely on rigid inductive biases, ignore inter-variable interactions, or apply static fusion strategies that limit adaptability across forecast horizons. These limitations create bottlenecks in capturing nuanced, horizon-specific relationships in time-series data. To solve this problem, we propose \textbf{T3Time}, a novel trimodal framework consisting of time, spectral, and prompt branches, where the dedicated frequency encoding branch captures the periodic structures along with a gating mechanism that learns prioritization between temporal and spectral features based on the prediction horizon. We also proposed a mechanism which adaptively aggregates multiple cross-modal alignment heads by dynamically weighting the importance of each head based on the features.
Extensive experiments on benchmark datasets demonstrate that our model consistently outperforms state-of-the-art baselines, achieving an average reduction of 3.28\% in MSE and 2.29\% in MAE. Furthermore, it shows strong generalization in few-shot learning settings: with 5\% training data, we see a reduction in MSE and MAE by 4.13\% and
1.91\%, respectively; and with 10\% data, by 3.62\% and 1.98\% on average. 
\end{abstract}

\begin{links}
    \link{Code}{https://github.com/monaf-chowdhury/T3Time/}
\end{links}

\section{Introduction}
Multivariate time-series forecasting (MTSF) lies at the heart of modern decision-making, powering everything from energy-load balancing \cite{liu2023sadi} and urban traffic management \cite{liu2024lighttr} to high-frequency trading \cite{xu2021anomaly} and weather forecasting \cite{schneider1974climate}. While the objective appears straightforward—predicting future values based on past observations—the underlying challenge is profoundly complex. Effective models must simultaneously capture short-term temporal fluctuations, long-range dependencies, and intricate inter-variable dynamics, all while maintaining computational efficiency and robustness in data-sparse regimes.

Recent advances in deep learning have led to the development of numerous models for time series forecasting \cite{miao2024unified,challu2023nhits}. However, early approaches were constrained by limited parameters and poor representations \cite{9665313}, hindering generalization. Following the development of transformer-based architectures \cite{vaswani2017attention}, frameworks like Informer \cite{zhou2021informer}, Autoformer \cite{wu2021autoformer}, and Fedformer \cite{zhou2022fedformer} addressed the quadratic complexity of standard self-attention and introduced trend-seasonal decomposition mechanisms.

Subsequent efforts have advanced in two complementary directions for better representation learning in the subspace of time-series. The first focuses on token restructuring, exemplified by PatchTST \cite{Yuqietal-2023-PatchTST}, which segments time series into sub-series patches and processes each channel independently. The second explores alternative representation spaces; for instance, Freeformer \cite{yue2025freeformer} operates in the frequency domain, demonstrating that spectral tokens can encode global periodic patterns more compactly than conventional time-domain attention. However, single-modality encoder architectures remain limited in their capacity to fully capture the intricate and multi-scale structure of temporal dynamics.

\begin{figure}[t]
    \centering
    \includegraphics[width=0.48\textwidth]{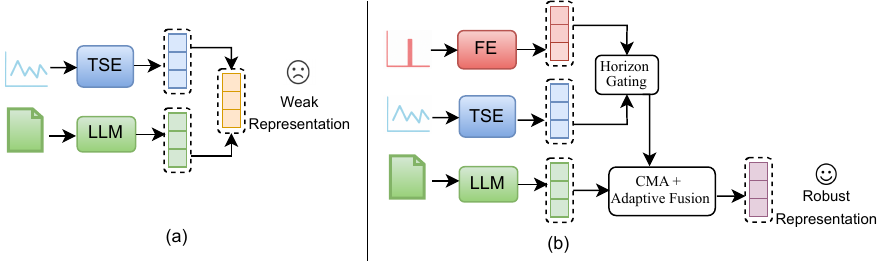}
    \caption{Comparison of bimodal vs. tri-modal framework for time series forecasting. (a) Bimodal models use static fusion of time and prompt features, lacking frequency awareness and horizon adaptivity. (b) T3Time introduces tri-modal encoding with horizon-aware gating and adaptive multi-head cross-modal alignment for robust, horizon-sensitive representations.}
    \label{fig:intro_plot}
\end{figure}

To make contextual time-series representations even robust, pre-trained LLM based \cite{radford2018improving} prompting techniques have been utilized in some frameworks. Prompts are either used to encapsulate the time-series information \cite{jia2024gpt4mts,huang2024leret} or to provide further context to the time series \cite{jin2023time,xue2023promptcast,xue2022leveraging}. Although time and prompt-based dual modal embeddings have accomplished promising performance, they have been plagued with embedding overlapping issues \cite{chang2024timedrl,jia2024gpt4mts}, which weakened representation integrity as seen here by Fig. \ref{fig:intro_plot}(a).  

Recently, TimeCMA \cite{liu2025timecma} has presented cross-modal alignment to integrate temporal-language dual modalities to represent embeddings as disentangled yet robust. Regardless, they miss out on aligning spectral modality to further represent global periodicity. By and large, existing MTSF frameworks exhibit three fundamental limitations. First, existing models often adopt a modality-isolated architecture, emphasizing a single modality while disregarding the other, which leads to fragmented representations. Second, if they utilize multi-modalities, they suffer from limited alignment capacity that constrain the model’s ability to capture rich, fine-grained interactions between multi modalities. Third, these models exhibit horizon rigidity by applying static processing across forecast lengths, ultimately hindering their ability to adapt modality emphasis based on the temporal scope of the prediction.

To address these challenges, we propose \textbf{T3Time} - illustrated in Fig. \ref{fig:intro_plot}(b), a tri-modal framework for MTSF that integrates temporal, spectral, and prompt-based semantic representations. Each modality captures complementary structures in the data. To enable effective multimodal integration, T3Time employs an adaptive multi-head cross-modal alignment module that dynamically weighs modality-specific embeddings based on their relevance to the forecasting task. 
Additionally, a horizon-aware gating mechanism modulates the influence of temporal and spectral features by prediction horizon, while a channel-wise residual fusion preserves variable-specific priors and enhances representational granularity. These components collectively enable T3Time to learn horizon-aware, variable-sensitive representations for robust generalization and improved forecasting. Our contributions are summarized as follows:

\begin{itemize}
\item We propose \textbf{T3Time}, a novel tri-modal forecasting framework that unifies temporal, spectral, and prompt-based semantic representations via an adaptive multi-head cross-modal alignment mechanism. This design enables dynamic, content-aware fusion of heterogeneous modalities for more expressive and context-aware representation.
\item We introduce horizon aware gating module and channel-wise residual fusion mechanism to elevate temporal adaptability and fine-grained feature representation.
\item T3Time consistently outperforms state-of-the-art baselines and demonstrates strong generalization across benchmark datasets.
\end{itemize}

\section{Related Work}

\textbf{Forecasting in Time Domain.} Time series forecasting has witnessed significant advances driven by deep learning \cite{lai2018modeling,franceschi2019unsupervised,jin2022adaptive}, with Transformer-based architectures \cite{wen2022transformers} emerging as powerful tools due to their capacity to model long-range dependencies. Early transformer-based methods \cite{zhou2021informer,wu2021autoformer,liu2022pyraformer}  introduce various attention mechanisms to reduce the quadratic complexity of standard self-attention and improve the efficiency of forecasting. However, most of them rely on fixed inductive biases (e.g., decomposition, sparse priors) and can not adapt to varying forecasting horizons. iTransformer \cite{liu2023itransformer} models each time series as an independent token, enabling flexible cross-variable attention. Regardless, its reliance on a single-stream encoder and static fusion limits its ability to capture complex, modality-specific dynamics in multivariate forecasting.

\textbf{Forecasting in Frequency Domain.} Frequency-domain methods \cite{cao2020spectral,woo2022cost,sun2022fredo,yi2023frequency,chen2023fraug} offer another perspective by modeling temporal periodicity and long-term structure. Autoformer \cite{wu2021autoformer} integrates series decomposition with autocorrelation in the frequency domain, while FEDformer \cite{zhou2022fedformer} further enhances this approach employing a mixture-of-expert structure and Fourier-based attention. Although effective, these methods typically employ fixed or global spectral representations, lacking mechanisms for adapting frequency-domain importance based on forecast length or contextual features.

\textbf{Cross-modal Alignment.} Recent work has explored adapting large language models (LLMs) \cite{lu2021fpt} for time series forecasting, either by replacing standard tokenizers with learned embeddings for time series inputs \cite{zhou2023one,liu2024spatial} or by formatting time series as textual prompts \cite{xue2023promptcast,jin2023time,pan2024s}. Although these methods leverage the representational power of LLMs, they often face modality mismatches. 
Moreover, fusion between language and numerical representations is typically static, limiting adaptability across tasks. 
Several studies have proposed the transfer of knowledge from pre-trained models using self-supervised learning \cite{zhang2022self,deldari2022beyond,zhang2024self}, multimodal reprogramming \cite{chen2024model}, or instruction-based fine-tuning \cite{yin2024survey}. TimeCMA \cite{liu2025timecma} represents a recent effort in this direction, introducing a cross-modal alignment to integrate time series and prompt embeddings for multivariate forecasting. 
However, its use of a single head alignment mechanism can be limiting, as it constrains the model's ability to capture diverse and fine-grained interactions between semantic and temporal signals. To this end, we introduce trimodal encodings to better represent time series representations and ensure robust performance.

\section{Methodology}
\label{sec:methodology}

\begin{figure*}[t]
    \centering
    \includegraphics[width=\textwidth]{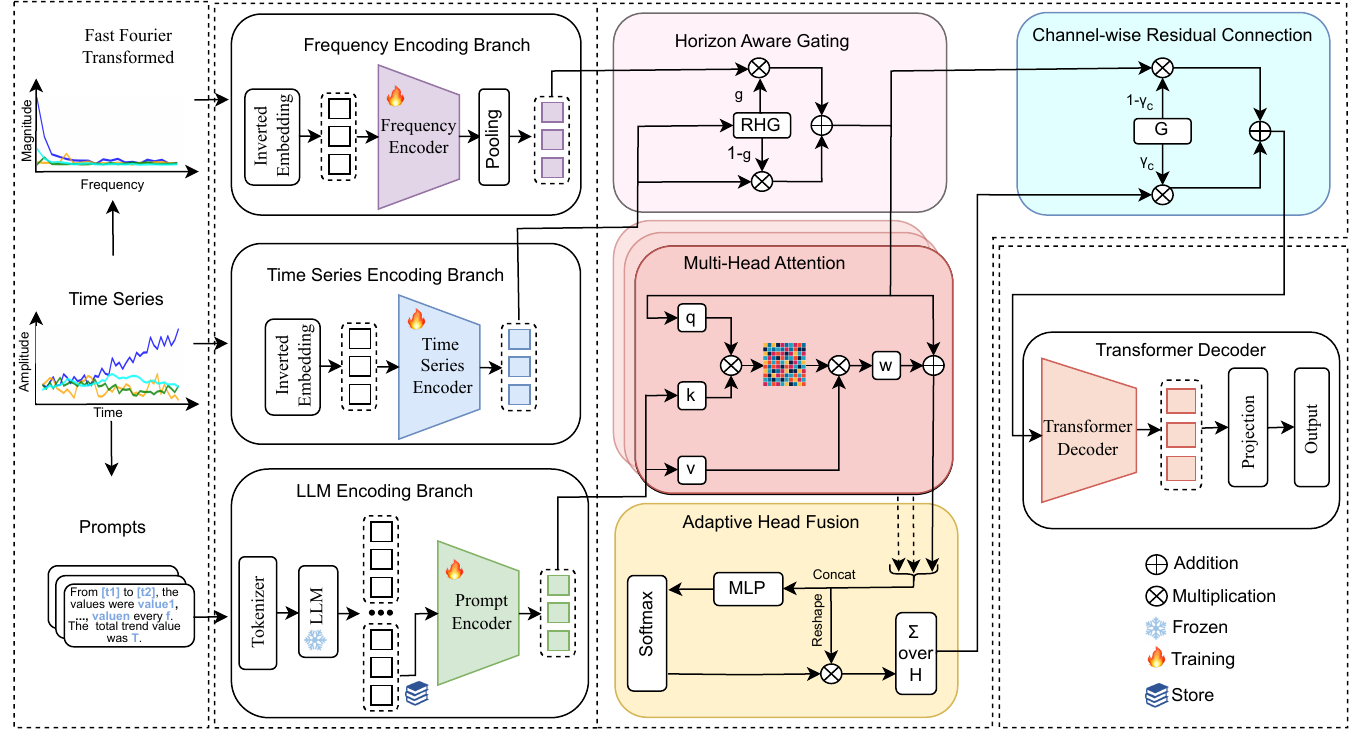}
    \caption{Overview of our framework. The model comprises tri-modal encoding (time, frequency, prompt), a horizon-aware gating module for dynamic temporal-spectral fusion, adaptive multi-head cross-modal alignment with per-head importance weighting, and a channel-wise residual connection for fine-grained representation mixing prior to decoding. }
    \label{fig:overview}
\end{figure*}

Our three-stage framework, as illustrated in Figure \ref{fig:overview} contains Tri-Modal Encoding, Adaptive Multi-Head Cross-Modal Alignment (CMA), and channel-wise residual connection. \textbf{Tri-Modal Encoding} comprises three branches: the frequency encoding branch, the time series encoding branch and the LLM encoding branch, designed to extract three different representations from the input time series. We also designed a representation-rich horizon-gating to fuse the frequency and time encoder branches dynamically based on the forecast horizon. \textbf{Adaptive Multi-head CMA}, inspired by \cite{vaswani2017attention}, aligns the fused
temporal-spectral features with the prompt embeddings using multiple instances of a Cross-Modal Attention module. Each attention head independently computes cross-attention between the fused temporal-spectral embeddings and prompt embeddings. Adaptive head fusion mechanism combines all the attention heads based on their importance and generates an expressive aligned representation. Finally, \textbf{Channel-wise Residual Connection} creates a fine-grained residual fusion between original temporal-spectral embeddings and aligned cross-modal representations, which are ultimately passed through the decoder for forecasting.

\subsection{Tri-Modal Encoding}
\indent\textbf{Frequency Encoding Branch:} To capture periodic and frequency-aware patterns from the time series, we designed a dedicated Frequency Encoding Branch, which transforms raw temporal inputs into the frequency domain using the real-valued fast Fourier transform and processes the resulting spectral features through a Transformer-based encoder.

Given the normalized input $\mathbf{X}_t \in \mathbb{R}^{B \times N \times L}$, where $B$ is the batch size, $N$ is the number of variables (or nodes), and $L$ is the sequence length, we apply the real-valued Fourier transform along the temporal dimension to obtain complex-valued spectra:
\begin{equation}
\widehat{\mathbf{X}}_t = \mathcal{F}_r(\mathbf{X}_t) \in \mathbb{C}^{B \times N \times L_f}, \quad L_f = \left\lfloor \frac{L}{2} \right\rfloor + 1
\end{equation}


Only the magnitude spectrum, $\mathbf{F}$, is retained as input to the frequency encoder. Each frequency bin is treated as a token. The tensor $\mathbf{F}$ is reshaped to $(BN) \times L_f$ and projected to the embedding dimension $C$ using a learnable projection matrix $\mathbf{W}_f \in \mathbb{R}^{C \times 1}$:
\begin{equation}
\mathbf{Z}_f = \phi\left( \mathbf{F} \cdot \mathbf{W}_f^\top \right) \in \mathbb{R}^{BN \times L_f \times C}
\end{equation}
where $\phi(\cdot)$ denotes an element-wise ReLU nonlinearity.

To model dependencies across frequency components, the projected frequency tokens are passed through a single-layer Transformer encoder. Let, $\mathcal{T}$ denote the self-attention block with pre-normalization. The encoded representation is given by:
\begin{equation}
\widetilde{\mathbf{Z}}_f = \mathcal{T}(\mathbf{Z}_f) \in \mathbb{R}^{BN \times L_f \times C}
\end{equation}

To aggregate the encoded spectral features, we compute a learnable attention-weighted pooling over the frequency bins to summarize the frequency information for each feature. Specifically, we first project the encoded frequency tokens \( \widetilde{\mathbf{Z}}_f \in \mathbb{R}^{BN \times L_f \times C} \) through a two-layer perceptron consisting of weights \( \mathbf{W}_1 \in \mathbb{R}^{C \times d} \) and \( \mathbf{W}_2 \in \mathbb{R}^{d \times 1} \), with an intermediate nonlinearity \( \phi(\cdot) \). The resulting scalar logits are normalized via a softmax function across the frequency dimension \( L_f \), yielding attention weights \( \boldsymbol{\alpha} \in \mathbb{R}^{BN \times L_f \times 1} \):

\begin{equation}
\boldsymbol{\alpha} = \frac{
\exp \left( \left[ \phi\left( \widetilde{\mathbf{Z}}_f \mathbf{W}_1 \right) \right] \mathbf{W}_2 \right)
}{
\sum_{j=1}^{L_f} \exp \left( \left[ \phi\left( \widetilde{\mathbf{Z}}_f \mathbf{W}_1 \right) \mathbf{W}_2 \right]_j \right)
}
\in \mathbb{R}^{BN \times L_f \times 1}
\end{equation}

The final pooled embedding is computed as a weighted sum:
\begin{equation}
\mathbf{F}_{\text{pooled}} = \sum_{l=1}^{L_f} \alpha_l \cdot \widetilde{\mathbf{Z}}_{f, :, l} \in \mathbb{R}^{BN \times C}
\end{equation}

Eventually, the pooled frequency features are reshaped back to match the node dimension, $\mathbf{\tilde{F}} \in \mathbb{R}^{B \times N \times C}$. This frequency-aware representation $\mathbf{\tilde{F}} \in \mathbb{R}^{B \times N \times C}$ captures periodic patterns and serves as one of the three modality-specific encodings for subsequent fusion.

\indent\textbf{Time Series Encoding Branch:} To model temporal dependencies and evolving patterns in raw time-domain signals, we construct a dedicated \textbf{Time Series Encoding Branch}. This branch transforms normalized time series into contextualized representations using a shared projection followed by a Transformer-based encoder.

Given the normalized input $\mathbf{X}_t \in \mathbb{R}^{B \times N \times L}$, we project the temporal dimension into a latent embedding space of dimension $C$ using a shared learnable projection matrix $\mathbf{W}_t \in \mathbb{R}^{L \times C}$:
\begin{equation}
\mathbf{Z}_t = \mathbf{X}_t \mathbf{W}_t \in \mathbb{R}^{B \times N \times C}
\end{equation}

This projection treats each node’s input sequence as a single vector, linearly embedding the temporal axis into $C$ features per node. To model interactions and dependencies across temporal patterns within each node, we apply a Transformer encoder with pre-normalization. Let $\mathcal{T}_t$ denote the time-domain Transformer encoder:
\begin{equation}
\widetilde{\mathbf{Z}}_t = \mathcal{T}_t(\mathbf{Z}_t) \in \mathbb{R}^{B \times N \times C}
\end{equation}

$\widetilde{\mathbf{Z}}_t$ provides positionally-aware, temporally contextualized embeddings for each node and serves as time modality-specific representation in our tri-modal framework.

\indent \textbf{LLM Encoding Branch:} To inject external priors and semantic structure into the forecasting model, we introduce a dedicated \textbf{LLM Encoding Branch}, which leverages a pre-trained frozen GPT-2 \cite{radford2019language} to encode prompt-based descriptions of input time series segments.

Given the normalized input series $\mathbf{X}_t \in \mathbb{R}^{B \times N \times L}$ and their associated temporal markers $\mathbf{M}_t \in \mathbb{R}^{B \times L \times D}$ (e.g., date, hour), we generate natural language prompts describing the sequence statistics. Each prompt is constructed for every feature. Prompt generation strategy is further discussed in Appendix B-1. 

Each prompt is tokenized using a GPT-2 tokenizer and fed into a pre-trained GPT-2 model. Let $\mathcal{L}(\cdot)$ denote the language model. For a given tokenized prompt $\mathbf{P}_{i,j}$ associated with the $j$-th node in the $i$-th sample, we obtain:

\begin{equation}
\mathbf{E}_{i,j} = \mathcal{L}(\mathbf{P}_{i,j}) \in \mathbb{R}^{T \times d_{\text{LLM}}}
\end{equation}
where $T$ is the number of tokens in the prompt and $d_{\text{LLM}}$ is the embedding dimension of the language model (e.g., $768$ for GPT-2). Since different prompts may produce variable-length token sequences, we uniformly pad the output with copies of the final token embedding to ensure consistent shape across all samples and features.

After batching, we extract the final token embedding per prompt as a compact summary:

\begin{equation}
\mathbf{Z}_{\text{LLM}}[i, j] = \mathbf{E}_{i,j}[-1] \in \mathbb{R}^{d_{\text{LLM}}}
\end{equation}

The final LLM-derived representation is a tensor $\mathbf{Z}_{\text{LLM}} \in \mathbb{R}^{B \times N \times d_{\text{LLM}}}$, which is projected and passed through a Transformer encoder before being used in downstream cross-modal fusion.

\indent We introduce a \textbf{Horizon-Aware Gating Module} to adaptively balance contributions from different modalities based on the forecast horizon. The core intuition is that short-term forecasts may benefit more from time-localized representations, whereas long-range forecasts can better leverage global periodic patterns captured in the frequency domain.

Time encoding, $\widetilde{\mathbf{Z}}_t \in \mathbb{R}^{B \times N \times C}$, is first pooled over the feature dimension to obtain a global summary per sample. We normalize the prediction length by a constant factor and concatenate it with the pooled time encoding in order to provide the forecast length as a continuous conditioning signal, $\mathbf{g}_{\text{in}} \in \mathbb{R}^{B \times (C+1)}$. The concatenated vector, $\mathbf{g}_{\text{in}}$, is processed through a lightweight two-layer MLP followed by a sigmoid nonlinearity to produce channel-wise gating weights:


\begin{equation}
\mathbf{g} = \sigma\left( \mathbf{W}_4 \cdot \phi(\mathbf{W}_3 \cdot \mathbf{g}_{\text{in}}^\top) \right)^\top \in \mathbb{R}^{B \times C}
\end{equation}
where $\mathbf{W}_3 \in \mathbb{R}^{(C+1) \times d}$, $\mathbf{W}_4 \in \mathbb{R}^{d \times C}$, $\sigma(\cdot)$ denotes the element-wise sigmoid nonlinearity.

The output of this module is a horizon-aware convex combination of the frequency and time representations:
\begin{equation}
\mathbf{Z}_{\text{g}} = \mathbf{g} \odot \mathbf{\tilde{F}} + (1 - \mathbf{g}) \odot \widetilde{\mathbf{Z}}_t \in \mathbb{R}^{B \times C \times N}
\end{equation}
where $\odot$ denotes element-wise multiplication broadcast over the feature dimension. This gating mechanism enables the model to adaptively shift focus between temporally localized and spectrally global features, as a function of both the input content and the desired forecast horizon.


\subsection{Adaptive Dynamic Head Cross-Modal Alignment}
To integrate heterogeneous contextual representations from the time-spectral and semantic domains, we follow the cross-modal alignment strategy introduced in \cite{liu2025timecma}, wherein the time series encoder output is aligned with the prompt encoder output through cross-attention. However, instead of relying on a single head cross-modal alignment as in the original approach, we extend this paradigm by introducing \emph{adaptive dynamic head fusion}, where multiple CMA heads are independently learned and their outputs are dynamically fused based on data-dependent gating scores.

Each CMA head $h$ maps the fused representation from the time and frequency branches, $\mathbf{Z}_{\text{g}} \in \mathbb{R}^{B \times C \times N}$, and prompt embeddings, $\mathbf{Z}_{\text{LLM}} \in \mathbb{R}^{B \times E \times N}$, into a head-specific aligned output $\mathbf{H}^{(h)} \in \mathbb{R}^{B \times C \times N}$ via an independent cross-attention mechanism with queries from \( \mathbf{Z}_{\text{g}} \) and keys/values from \( \mathbf{Z}_{\text{LLM}} \). Rather than aggregating these heads through static averaging or fixed linear projections, head-wise outputs are aggregated via a \emph{feature-aware, head-adaptive} fusion to dynamically weight the importance of each head per feature. Necessarily, the outputs are first concatenated along the channel dimension and transposed to shape $\mathbb{R}^{B \times N \times HC}$:
\begin{equation}
\mathbf{U} = \left[ \mathbf{H}^{(1)}; \dots; \mathbf{H}^{(H)} \right]^\top \in \mathbb{R}^{B \times N \times HC}
\end{equation}

Each node's fused embedding vector $\mathbf{U}_{b,n} \in \mathbb{R}^{HC}$ (for batch index $b$ and node index $n$) is passed through a two-layer gating network to compute importance scores $\boldsymbol{\pi}_{b,n} \in \mathbb{R}^{H}$ over the CMA heads:

\begin{align}
\mathbf{e}_{b,n} &= \mathbf{W}_6 \, \phi\left( \text{LN}\left( \mathbf{W}_5 \mathbf{U}_{b,n}^\top \right) \right) \in \mathbb{R}^{H}, \\
\pi_{b,n}^{(h)} &= \frac{\exp(e_{b,n}^{(h)})}{\sum_{j=1}^{H} \exp(e_{b,n}^{(j)})}, \quad \sum_{h=1}^H \pi_{b,n}^{(h)} = 1
\end{align}
where $\mathbf{W}_5 \in \mathbb{R}^{128 \times HC}$ and $\mathbf{W}_6 \in \mathbb{R}^{H \times 128}$ are learnable matrices, $\phi(\cdot)$ is a pointwise ReLU nonlinearity, and $\text{LN}(\cdot)$ denotes layer normalization.

Let $\mathbf{H}^{(h)}_{b,:,n} \in \mathbb{R}^{C}$ be the $h$-th head output for sample $b$ and node $n$. The gated fusion is obtained via a convex combination over the head dimension, weighted by the attention scores $\pi_{b,n}^{(h)}$:
\begin{equation}
\boldsymbol{\Lambda}_{b,:,n} = \sum_{h=1}^{H} \pi_{b,n}^{(h)} \cdot \mathbf{H}^{(h)}_{b,:,n} \in \mathbb{R}^{C}
\end{equation}
yielding the final cross-modally aligned representation $\boldsymbol{\Lambda} \in \mathbb{R}^{B \times C \times N}$ across all $B$ samples and $N$ features.

\subsection{Channel-wise Residual Connection}
Before decoding, we apply a channel-wise residual fusion to reconcile horizon-aware spectral-temporal features with cross-modal alignment outputs, allowing each latent feature to balance its dependence on intrinsic patterns and external priors.

Here, $\mathbf{Z}_{\text{g}} \in \mathbb{R}^{B \times C \times N}$ denotes the horizon-gated fusion of time-domain and frequency-domain encodings, and $\boldsymbol{\Lambda} \in \mathbb{R}^{B \times C \times N}$ denotes the output of the adaptive multi-head cross-modal alignment. We learn a set of trainable channel-wise residual coefficients $\boldsymbol{\gamma} \in \mathbb{R}^{C}$ that modulate the importance of the two streams during the fusion process.

The fused representation $\boldsymbol{\Theta} \in \mathbb{R}^{B \times C \times N}$ is computed as a convex combination along the channel axis:
\begin{equation}
\boldsymbol{\Theta}_{b, c, n} = \boldsymbol{\gamma_c} \odot \boldsymbol{\Lambda}_{b, c, n} + (1 - \boldsymbol{\gamma_c}) \odot \mathbf{Z}_{\text{g}, b, c, n}
\end{equation}
where \( \odot \) denotes element-wise multiplication, $b$, $c$, and $n$ index the batch, channel, and node dimensions respectively, and $\boldsymbol{\gamma_c} \in [0, 1]$ is a learnable scalar specific to channel $c$.

This $\boldsymbol{\Theta} \in \mathbb{R}^{B \times C \times N}$ formulation allows each latent dimension to adaptively balance cross-modal information against temporal and spectral evidence, enabling fine-grained control over representational mixing. The fused representation $\boldsymbol{\Theta}$ is subsequently passed into the decoder for final forecasting. \\

\indent \textbf{Decoder:} Fused representation $\boldsymbol{\Theta} \in \mathbb{R}^{B \times C \times N}$, which encapsulates spectro-temporal priors and semantically grounded alignment, is transposed to match the input format of the transformer decoder. The decoder module, composed of $D$ stacked multi-head cross-attention blocks with pre-normalization, integrates global dependencies across nodes and channels. 

Let $\mathcal{D}(\cdot)$ denote the Transformer decoder. The contextualized output $\mathbf{Z}_d \in \mathbb{R}^{B \times N \times C}$ is given by:
\begin{equation}
\mathbf{Z}_d = \mathcal{D}\left( \boldsymbol{\Theta}^{\top}, \boldsymbol{\Theta}^{\top} \right), \quad \boldsymbol{\Theta}^{\top} \in \mathbb{R}^{B \times N \times C}
\end{equation}

Following decoding, the representation $\mathbf{Z}_d$ is linearly projected along the channel axis to produce the final forecast sequence of length $L_p$:
\begin{equation}
\widehat{\mathbf{Y}} = \mathbf{Z}_d \, \mathbf{W}_p^\top + \mathbf{b}_p, \quad \mathbf{W}_p \in \mathbb{R}^{L_p \times C}, \quad \widehat{\mathbf{Y}} \in \mathbb{R}^{B \times N \times L_p}
\end{equation}

Forecast tensor $\widehat{\mathbf{Y}}$ is transposed to match the original input format $\mathbf{Y} \in \mathbb{R}^{B \times L_p \times N} $, representing the final output. \\

\section{Experiments}
\label{sec:experiments}

T3Time demonstrates consistent and superior performance across a wide range of benchmarks, particularly excelling in long-horizon forecasting and low-data regimes. To ensure fair comparison, we adhere strictly to the experimental protocol established by \citet{liu2025timecma} across all baseline models unless explicitly stated otherwise. Our evaluation includes a comprehensive set of strong baselines: \textit{prompt-based LLMs} such as TimeCMA~\cite{liu2025timecma}, Time-LLM~\cite{jin2023time}, and UniTime~\cite{liu2024unitime}; \textit{Transformer-based architectures} including iTransformer~\cite{liu2023itransformer}, OFA (GPT4TS)~\cite{zhou2023one}, PatchTST~\cite{Yuqietal-2023-PatchTST}, and Fedformer~\cite{zhou2022fedformer}; the \textit{linear} model DLinear~\cite{zeng2023transformers}; and the \textit{CNN-based} framework TimesNet~\cite{wu2022timesnet}. Details regarding device, datasets, and evaluation metrics are provided in Appendix~A.

\begin{table*}[ht]
\centering
\scriptsize 
\setlength{\tabcolsep}{4.5pt}
\renewcommand{\arraystretch}{0.8}

\begin{tabular}{c|cc|cc|cc|cc|cc|cc|cc|cc|cc}
\toprule
\multirow{2}{*}{\textbf{Dataset}}  & 
\multicolumn{2}{c|}{\textbf{Ours}} & 
\multicolumn{2}{c|}{\textbf{TimeCMA}} & 
\multicolumn{2}{c|}{\textbf{TimeLLM}} & 
\multicolumn{2}{c|}{\textbf{UniTime}} & 
\multicolumn{2}{c|}{\textbf{TimesNet}} & 
\multicolumn{2}{c|}{\textbf{DLinear}} & 
\multicolumn{2}{c|}{\textbf{iTransformer}} & 
\multicolumn{2}{c|}{\textbf{PatchTST}} & 
\multicolumn{2}{c}{\textbf{OFA}} \\
 & MSE & MAE & MSE & MAE & MSE & MAE & MSE & MAE & MSE & MAE & MSE & MAE & MSE & MAE & MSE & MAE& MSE & MAE \\
\midrule

{ETTm1} 
&  \textbf{{0.372}}&\underline{0.393}  &  \underline{0.380}&\textbf{{0.392}} & 0.410&0.409 & 0.385&0.399 & 0.400&0.406 & 0.403&0.407 & 0.407&0.410 & 0.392&0.402 & 0.396&0.401 \\
\midrule

{ETTm2} 
&  \underline{0.279}&\textbf{{0.322}}  &  \textbf{{0.275}}&\textbf{{0.323}} &  0.296&0.340 & 0.293&0.334 &  0.291&0.333 & 0.350&0.401 & 0.288&0.332 & 0.285&\underline{0.328} & 0.294&0.339\\
\midrule

{ETTh1} 
& \textbf{{0.418}}&\textbf{{0.430}}  &  \underline{0.423}&\underline{0.431} & 0.448&0.443 & 0.442&0.448 & 0.458&0.450 & 0.456&0.452 & 0.454&0.447 & 0.463&0.449 & 0.457&0.450 \\
\midrule

{ETTh2} 
& \textbf{{0.348}}&\textbf{{0.390}} & \underline{0.372}&\underline{0.397} & 0.381&0.404 & 0.378&0.403 & 0.414&0.427 & 0.559&0.515 & 0.383&0.407 & 0.395&0.414 & 0.389&0.414 \\
\midrule

{ECL}  
& \textbf{{0.170}}&\textbf{{0.266}} & 0.174&0.269 & 0.195&0.288 & 0.216&0.306 &  0.192&0.295 & 0.212&0.300 & 0.178&0.270 & 0.207&0.289 & 0.217&0.308\\
\midrule

{Weather}  & \textbf{{0.244}}&\textbf{{0.275}} & \underline{0.250}&\underline{0.276} & 0.275&0.291 & 0.253&\underline{0.276} & 0.259&0.287 & 0.265&0.317 & 0.258&0.278 & 0.257&0.280 & 0.279&0.297 \\
\midrule

{ILI} 
& \textbf{{1.705}}&\textbf{{0.835}} & \underline{1.922}&\underline{0.921} & 2.432&1.012 & 2.108&0.929 &  2.139&0.931 & 2.616&1.090 & 2.444&1.203 & 2.388&1.011 & 2.623&1.060 \\
\midrule

{Exchange}  & \textbf{{0.353}}&\textbf{{0.401}} & 0.395 & 0.429 & 0.372 & 0.416 & 0.364 & 0.404 &  0.416& 0.443 & \underline{0.354} & 0.414 & 0.360 & \underline{0.403} & 0.390 & 0.429 & 0.519 & 0.500 \\
\midrule

\textbf{1st Count} &  \textbf{{7}}&\textbf{{7}}  &  \underline{1}&\underline{2} &  0&0  &  0&0  &  0&0  &  0&0  &  0&0  &  0&0  &  0&0  \\

\bottomrule
\end{tabular}

\caption{Multivariate forecasting results. All results are averaged from four different forecasting horizons: $\mathbf{H} \in \{96, 192, 336, 720\}$ for the input sequence length $96$. \textbf{Bold}: the best, \underline{underline}: the second best. Full results are in Appendix~C.}
\label{tab:multivariate_results}
\end{table*}

\begin{table*}[!ht]
\centering
\scriptsize
\setlength{\tabcolsep}{4.5pt}
\renewcommand{\arraystretch}{0.8}
\begin{tabular}{c|cc|cc|cc|cc|cc|cc|cc|cc}
\toprule
\multirow{2}{*}{\textbf{Dataset}} & 
\multicolumn{2}{c|}{\textbf{Ours}} & 
\multicolumn{2}{c|}{\textbf{TimeCMA}} & 
\multicolumn{2}{c|}{\textbf{TimeLLM}} & 
\multicolumn{2}{c|}{\textbf{GPT4TS}} & 
\multicolumn{2}{c|}{\textbf{TimesNet}} & 
\multicolumn{2}{c|}{\textbf{DLinear}} & 
\multicolumn{2}{c|}{\textbf{PatchTST}} & 
\multicolumn{2}{c}{\textbf{Fedformer}} \\
& MSE & MAE & MSE & MAE & MSE & MAE & MSE & MAE & MSE & MAE & MSE & MAE & MSE & MAE & MSE & MAE \\
\midrule

ETTm1   & \textbf{0.376}&\textbf{0.398} & \underline{0.387}&\underline{0.410} & 0.404&0.427 & 0.464&0.441 & 0.677&0.537 & 0.411&0.429 & 0.501&0.466 & 0.722&0.605\\
\midrule

ETTm2   &  \textbf{0.266}&\underline{0.327} & 0.312&0.358 & \underline{0.277}&\textbf{0.323} & 0.293&0.335 & 0.320&0.353 & 0.316&0.368 & 0.296&0.343 & 0.463&0.488\\
\midrule

ETTh1  & \textbf{0.449}&\textbf{0.454} & \underline{0.480}&\underline{0.479} & 0.556&0.522 & 0.590&0.525 & 0.869&0.628 & 0.691&0.600 & 0.633&0.542 & 0.639&0.561\\
\midrule

ETTh2   & \textbf{0.357}&\textbf{0.388} & 0.398&0.433 & \underline{0.370}&\underline{0.394} & 0.397& 0.421 & 0.479&0.465 & 0.605&0.538 & 0.415&0.431 & 0.466&0.475\\
\midrule

Weather & \textbf{0.226}&\textbf{0.268} & \underline{0.229}&\underline{0.272} & 0.234&0.273 & 0.238&0.275 & 0.279&0.301 & 0.241&0.283 & 0.242&0.279 & 0.284&0.324\\

\midrule

\textbf{1st Count}
  &   \textbf{5}&\textbf{4}  &  0 & 0  &  0&\underline{1} &  0&0  &  0&0  &  0&0  &  0&0 & 0&0  \\

\bottomrule
\end{tabular}
\caption{Few-shot learning on 10\% training data. All results are averaged from four different forecasting horizons: $\mathbf{H} \in \{96, 192, 336, 720\}$ for the input sequence length $512$. \textbf{Bold}: the best, \underline{underline}: the second best. Full results are in Appendix~D.}
\label{tab:few_shot_10}
\end{table*}

\begin{table*}[!ht]
\centering
\scriptsize
\setlength{\tabcolsep}{4.5pt}
\renewcommand{\arraystretch}{0.8}
\begin{tabular}{c|cc|cc|cc|cc|cc|cc|cc|cc}
\toprule
\multirow{2}{*}{\textbf{Dataset}} & 
\multicolumn{2}{c|}{\textbf{Ours}} & 
\multicolumn{2}{c|}{\textbf{TimeCMA}} & 
\multicolumn{2}{c|}{\textbf{TimeLLM}} & 
\multicolumn{2}{c|}{\textbf{GPT4TS}} & 
\multicolumn{2}{c|}{\textbf{TimesNet}} & 
\multicolumn{2}{c|}{\textbf{DLinear}} & 
\multicolumn{2}{c|}{\textbf{PatchTST}} & 
\multicolumn{2}{c}{\textbf{Fedformer}} \\
& MSE & MAE & MSE & MAE & MSE & MAE & MSE & MAE & MSE & MAE & MSE & MAE & MSE & MAE & MSE & MAE \\
\midrule
	
ETTm1   & \textbf{{0.384}}&\textbf{{0.405}} & \underline{0.396}&\underline{0.416} & 0.425&0.434 & 0.472&0.450 & 0.717&0.561 & 0.400&0.417 & 0.526&0.476 & 0.730&0.592\\
\midrule

ETTm2   &    \textbf{{0.267}}&\textbf{{0.330}} & 0.329&0.367 & \underline{0.274}&\underline{0.323} & 0.308&0.346 & 0.344&0.372 & 0.399&0.426 & 0.314&0.352 & 0.381&0.404\\
\midrule

ETTh1  & \textbf{{0.442}}&\textbf{{0.451}} & \underline{0.472}&\underline{0.470} & 0.627&0.543 & 0.681&0.560 & 0.925&0.647 & 0.750&0.611 & 0.694&0.569 & 0.658&0.562\\
\midrule

ETTh2   & \textbf{{0.357}}&\textbf{{0.403}} & 0.395&0.430 & \underline{0.382}&\underline{0.418} & 0.400&0.433 & 0.439&0.448 & 0.694&0.577 & 0.827&0.615 & 0.463&0.454\\
\midrule
	
Weather & \textbf{{0.226}}&\textbf{{0.269}} & \underline{0.231}&\underline{0.273} & 0.260&0.309 & 0.263&0.301 & 0.298&0.318 & 0.263&0.308 & 0.269&0.303 & 0.309&0.353\\

\midrule

\textbf{1st Count}
  & \textbf{{5}}&\textbf{{5}}  &  0&0  &  0&0 &  0&0  &  0&0  &  0&0  &  0&0 & 0&0  \\

\bottomrule
\end{tabular}
\caption{Few-shot learning on 5\% training data. All results are averaged from four different forecasting horizons: $\mathbf{H} \in \{96, 192, 336, 720\}$ for the input sequence length $512$. \textbf{Bold}: the best, \underline{underline}: the second best. Full results are in Appendix~D.}
\label{tab:few_shot_5}
\end{table*}

\subsection{Long-Term Forecasting}
\textbf{Setups.} We evaluate T3Time on eight widely-used multivariate time series benchmarks: ETTh1, ETTh2, ETTm1, ETTm2, ECL, Weather, ILI, and Exchange. Following the standardized protocol from \citet{liu2025timecma}, we set the input sequence length to $96$ and vary the forecasting horizon across $\{96, 192, 336, 720\}$, except for ILI where we follow the $\{24, 36, 48,60\}$ prediction steps. Mean squared error (MSE) and mean average error (MAE) are used as evaluation metrics for all our experiments.


\textbf{Results.} Table~\ref{tab:multivariate_results} summarizes the forecasting performance across all datasets. On average, T3Time consistently achieves state-of-the-art (SOTA) results in $14$ out of $16$ baselines. 
The model yields the lowest error across most horizons, outperforming recent competitive approaches such as TimeLLM, ~\cite{jin2023time} and iTransformer, ~\cite{liu2023itransformer} by 11.28\% in MSE and 6.20\% in MAE, and 8.86\% in MSE and 6.10\% in MAE, respectively. When compared to the strongest prompt-based model, TimeCMA, ~\cite{liu2025timecma}, T3Time reduces the MSE by up to 4.36\% on average, while demonstrating more stable MAE performance across long horizons. Overall, T3Time achieves an average MSE reduction of 3.28\% and an MAE reduction of 2.29\% compared to the SOTA baselines. A comprehensive overview of our full results is provided in Appendix~C.


\begin{table*}[!ht]
\centering
\scriptsize
\setlength{\tabcolsep}{4.5pt}
\renewcommand{\arraystretch}{0.8}
\begin{tabular}{c|cc|cc|cc|cc|cc|cc|cc|c}
\toprule
\multirow{2}{*}{\textbf{Design}} & 
\multicolumn{2}{c|}{\textbf{ETTm1}} & 
\multicolumn{2}{c|}{\textbf{ETTm2}} & 
\multicolumn{2}{c|}{\textbf{ETTh1}} & 
\multicolumn{2}{c|}{\textbf{ETTh2}} & 
\multicolumn{2}{c|}{\textbf{Weather}} & 
\multicolumn{2}{c|}{\textbf{ILI}} & 
\multicolumn{2}{c|}{\textbf{Exchange Rate}} & 
\multirow{2}{*}{\textbf{1st Count}} \\ 
& MSE & MAE & MSE & MAE & MSE & MAE & MSE & MAE & MSE & MAE & MSE & MAE & MSE & MAE & \\ 

\midrule
						
Our Model   & \textbf{{0.372}}&\textbf{{0.393}}  & \textbf{{0.279}}	&\textbf{{0.322}}	&  \textbf{{0.418}}	&\textbf{{0.430}}	& \textbf{{0.348}}	&\textbf{{0.390}}	& \textbf{{0.244}}	&\textbf{{0.275}}  &  \textbf{{1.705}}	&\textbf{{0.835}}	&\textbf{{0.353}}	&\textbf{{0.401}} 
& \textbf{14}\\
\midrule

w/o Frequency Module   & 0.381&0.394 & \textbf{{0.279}}&0.324 & 0.433&0.433 & 0.364&0.398 & 0.250&0.280 & 1.786 & 0.884 & 0.374	&0.410  & 1\\
\midrule

w/o Multihead CMA  & 0.374&0.395 & 0.283&0.325 & 0.421&0.432 & 0.355&0.392 & 0.249&0.277 & 1.813&0.865	& 0.378&0.413 & 0\\
\midrule

w/o Residual Connection   &  0.404&0.413 & 0.288&0.329	& 0.433&0.443 &  0.384&0.411 & 0.249&0.280	& 2.176&0.969 & 0.396&0.427 & 0\\
\midrule
	
w/o Gating Mechanism & 0.373&0.396 &0.280	&\textbf{{0.322}} &0.425  &0.432 &0.363	&0.398 &0.249	& 0.277  & 1.724&0.837 & 0.373&0.412 & 1\\

\bottomrule
\end{tabular}
\caption{Results of design choices related to the model. All results are averaged from four different forecasting horizons: $\mathbf{H} \in \{96, 192, 336, 720\}$ for the input sequence length $96$. \textbf{Bold}: the best. Full results are in Appendix~E. }
\label{tab:design_variant}
\end{table*}

\subsection{Few-Shot Forecasting}
\textbf{Setups.} Over the years the foundation models have shown exceptional performance in generalization tasks by few-shot learning or zero-shot learning settings \cite{brown2020language,achiam2023gpt}. To evaluate the generalization performance of T3Time under few-shot conditions, we follow the setups from \citet{jin2023time} for a fair comparison. Specifically, we adopt a scenario where the training data consists of $10\%$ and $5\%$ of the available time steps, with input sequence length set to $512$. We evaluate T3Time on the same benchmarks used in the long-term forecasting experiments.

\textbf{Results.} Tables~\ref{tab:few_shot_10} and~\ref{tab:few_shot_5} summarize the results for $10\%$ and $5\%$ few-shot learning, respectively. In both settings, T3Time outperforms almost all SOTA baselines. Specifically, for the recent SOTA models such as TimeCMA, TimeLLM, and GPT4TS (OFA) in the 10\% few-shot task, average MSE is reduced by 7.13\%, 7.42\%, and 13.44\%, respectively. T3Time reduces average MSE and MAE by 3.62\% and 1.98\% for 10\% few-shot forecasting tasks in all SOTA scores. Similar trend is visible in the 5\% few-shot forecasting task, as T3Time improves upon SOTA scores by 4.13\% and 1.91\% w.r.t. MSE and MAE. When compared with models like TimesNet, DLinear, and PatchTST, T3Time consistently outperforms these approaches, surpassing $30\%$ in MSE reduction. Additionally, relative to recent SOTA methods like TimeCMA, TimeLLM, and GPT4TS, T3Time demonstrates an average MSE improvement exceeding $10\%$. A comprehensive overview of our full results for few shot forecasting is provided in Appendix~D.


\subsection{Design Variant}
\textbf{Setups.} To evaluate the robustness and contribution of various design choices in T3Time, we experiment with several ablations, including removing frequency features, multi-head mechanisms, residual connections, and gating. Specifically, we evaluate the following design variants: (1) \textit{Without Frequency}, where frequency-domain features are excluded; (2) \textit{Without Multi-Head}, where the adaptive multi-head cross-modal alignment is disabled; (3) \textit{Without Residual}, which eliminates the channel-wise residual fusion mechanism; and (4) \textit{Without Gating}, where the horizon-aware gating mechanism is omitted. We maintain the same setup as the long-term forecasting.

\textbf{Results.} Table~\ref{tab:design_variant} presents the performance comparison across different design variants. The full model achieves average SOTA performance in 14 out of 14 cases for both MSE and MAE. Notably, removing frequency-domain features results in a 3.22\% MSE and 1.85\% MAE drop compared to the full model. This emphasizes the crucial role of frequency-domain information in capturing both temporal and spectral patterns. However, the residual connection's contribution is the most significant as excluding it leads to the largest performance drop, with an average MSE increase of 8.36\% and MAE of 5.25\%. Excluding the gating mechanism and multi-head CMA results in a more modest decline, with an average MSE decrease of approximately 2\%. Full results for the design variants are provided in Appendix~E.

\subsection{t-SNE Visualization}
\begin{figure}[!h]
    \centering
    \includegraphics[width=0.45\textwidth, height=5cm]{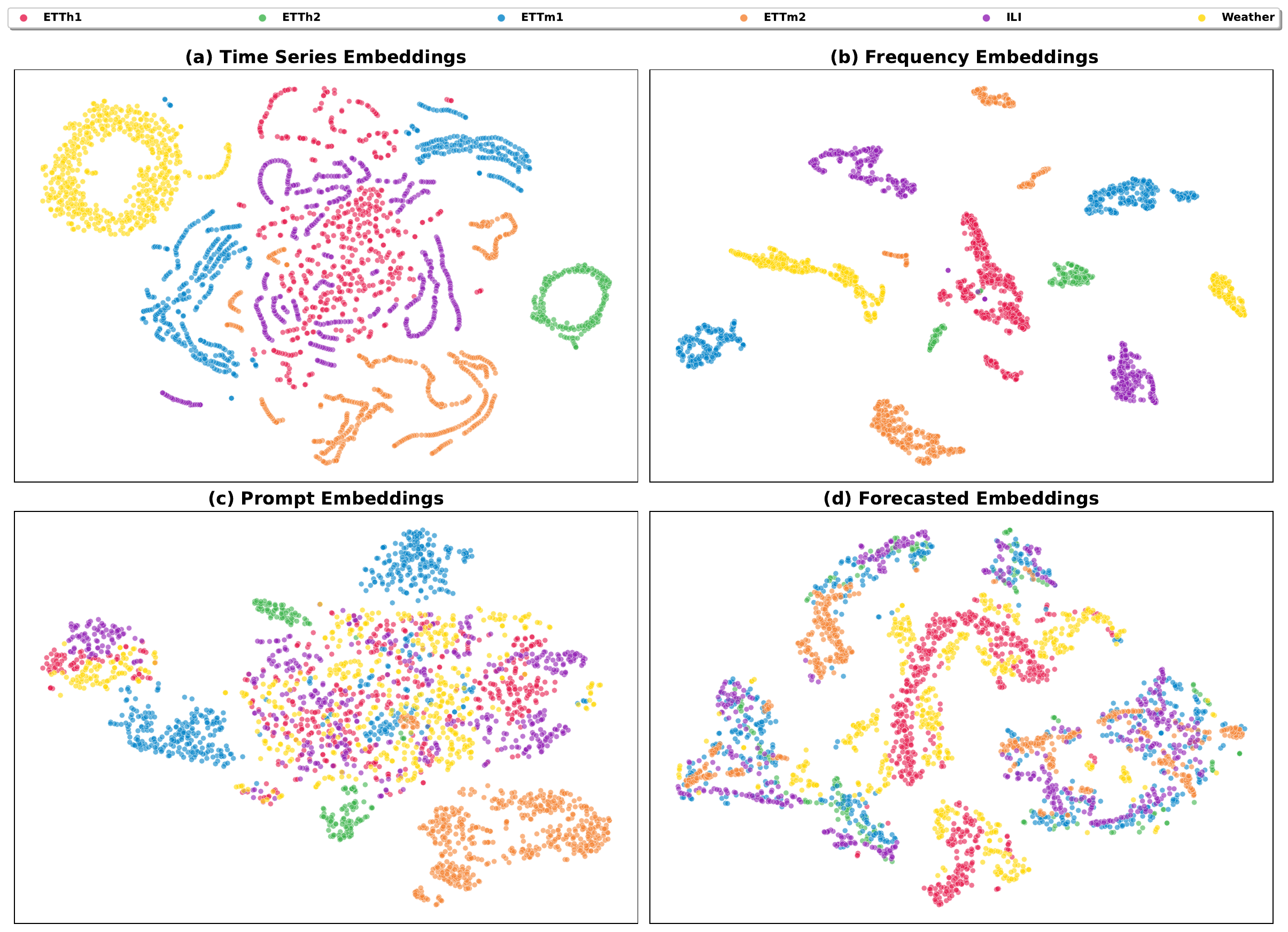}
    \caption{The combined T-SNE visualization of time series, frequency, prompt, and forecasted embeddings across six datasets. A detailed breakdown of each modality and dataset is provided in Appendix F.}
    \label{fig:tsne-combined}
\end{figure}
To better understand the effectiveness of the learned embeddings, we employ T-SNE visualization. Figure~\ref{fig:tsne-combined} presents the T-SNE visualization of the embeddings across four key modalities: time series embeddings, frequency embeddings, prompt embeddings, and forecasted embeddings. Panel \ref{fig:tsne-combined}(a) illustrates the time series embeddings, which demonstrate clear clustering of the datasets. Separation between the datasets indicates that the model is effectively capturing their temporal characteristics. Panel \ref{fig:tsne-combined}(b) visualizes the frequency embeddings, where we observe a distinct clustering pattern. This suggests that the model captures meaningful frequency-domain information, which is crucial for distinguishing underlying periodicities in the time series. Panel \ref{fig:tsne-combined}(c) shows the prompt embeddings, which exhibit a relatively more dispersed structure compared to the time series and frequency embeddings, reflecting the diverse nature of the prompts. Finally, panel \ref{fig:tsne-combined}(d) displays the forecasted embeddings, where the clustering is similar to that of the time series embeddings, suggesting that the model is learning representations that align with both historical and forecasted data. These visualizations illustrate T3Time's ability to integrate temporal, spectral, and prompt information in learning robust multimodal representation.

\section{Conclusion}
In this work, we present \textbf{T3Time}, a novel framework for multivariate time series forecasting which integrates three distinct modalities—time series, frequency, and prompt embeddings. Along with the tri-modal encoding framework, T3Time adapts the fusion of temporal-spectral features using a dynamic horizon-aware gating mechanism. The introduction of adaptive multi-head cross-modal alignment enables more flexible and contextual interaction between the time-spectral and prompt representations, while the channel-wise residual connection ensures efficient fusion of these representations later in the framework. T3Time consistently outperforms SOTA models, achieving notable reductions in both MSE and MAE across numerous baselines. Additionally, T3Time demonstrates strong generalization in few-shot learning scenarios, with significant performance gains. Future works can explore large-scale pertaining and better representation methods to enrich modalities for stronger time series forecasting.

\bibliography{aaai2026} 
\newpage

\section{Appendix A: Experimental Setup}
\subsection {A.1: Hardware Details} 
All experiments, including the ablation studies, were conducted on a workstation equipped with an Intel Core i9-285K processor (24 cores, 3.7 GHz base clock), 96 GB of RAM, and an NVIDIA GeForce RTX 4090 GPU with 24 GB of dedicated VRAM. The system ran on Microsoft Windows 11 Pro (build 26100) with DirectX $12$ and CUDA $12$ support. The motherboard used was MSI MPG Z690 Carbon WiFi, and the GPU driver version was $32.0.15.6614$. All code was executed in a Python $3.11$, PyTorch $2.1.2$, torchvision $0.8.0$, and conda environment with PyTorch \cite{paszke2019pytorch} and Transformers frameworks. All of the experiments were done using three different seed values, and the average score was reported.

\begin{table*}[ht]
\centering
\scriptsize
\setlength{\tabcolsep}{4.5pt}
\renewcommand{\arraystretch}{0.8}
\begin{tabular}{c|c|c|c|c|c}
\toprule
\textbf{Dataset} & \textbf{Dimension} & \textbf{Series Length} & \textbf{Dataset Size} & \textbf{Frequency} & \textbf{Domain} \\
\midrule
ETTm1 & 7 & $\{96, 192, 336, 720\}$ & $(34465, 11521, 11521)$ & 15 mins & Electricity \\ \midrule
ETTm2 & 7 & $\{96, 192, 336, 720\}$ & $(34465, 11521, 11521)$ & 15 mins & Electricity \\ \midrule
ETTh1 & 7 & $\{96, 192, 336, 720\}$ & $(8545, 2881, 2881)$ & 15 mins & Electricity \\
\midrule
ETTh2 & 7 & $\{96, 192, 336, 720\}$ & $(8545, 2881, 2881)$ & 15 mins & Electricity \\
\midrule
Electricity (ECL) & 321 & $\{96, 192, 336, 720\}$ & $(18317, 2633, 5261)$ & Hourly & Electricity \\  \midrule
Weather & 21 & $\{96, 192, 336, 720\}$ & $(36792, 5271, 10540)$ & 10 mins & Weather \\
\midrule
Exchange & 8 & $\{96, 192, 336, 720\}$ & $(5120, 665, 1422)$ & Daily & Exchange rate \\
\midrule
ILI &  7 & $\{24, 36, 48, 60\}$ & $(617, 74, 170)$ & Weekly & Illness \\
\bottomrule
\end{tabular}
\caption{Summary of datasets used in our experiments.
Each dataset varies in domain, dimensionality, sampling frequency, and series length. Forecasting is performed over multiple horizons, with input sequence length fixed to $96$ (except ILI). Dataset size is organized as a train set, a validation set, and a test set. Dimensions describe the number of time series channels.}
\label{tab:datasets}
\end{table*}

\subsection {A.2: Dataset Details}
\label{appendix:dataset_details}
Overall dataset statistics are illustrated in Table \ref{tab:datasets}. We evaluate the performance of time series forecasting on seven widely used benchmark datasets, including ETT (ETTm1, ETTm2, ETTh1, ETTh2) \cite{zhou2021informer}, ECL \cite{asuncion2007uci}, Weather \cite{wu2021autoformer}, Exchange \cite{wu2022timesnet}, and ILI \cite{wu2022timesnet}. 

\textbf{ETT.} The Electricity Transformer Temperature (ETT) dataset serves as a critical benchmark for evaluating electric power forecasting. It comprises two years of data collected from two separate counties in China. To analyze the impact of temporal granularity, the dataset is divided into four subsets with different sampling frequencies: ETTh1 and ETTh2 are sampled at 1-hour intervals, while ETTm1 and ETTm2 are sampled at 15-minute intervals. Each data point contains six power load-related features along with a target variable, oil temperature.

\textbf{ECL.} The Electricity dataset includes hourly electricity consumption data from 370 clients, providing insights into consumer-level load patterns. Data is collected from 1st January, 2011 with a sampling interval of 15 minutes. 

\textbf{Weather.} Weather dataset consists of one year of meteorological measurements recorded every 10 minutes across 21 weather stations of the Max Planck Biogeochemistry Institute in Germany. It includes 21 variables such as air temperature, humidity, and wind speed, etc.

\textbf{Exchange.} Exchange comprises daily exchange rate records from 1990 to 2016 for eight foreign currencies, including those of Australia, the United Kingdom, China, Japan, Canada, Singapore, Switzerland, and New Zealand. The data is sampled at a one-day interval.

\textbf{ILI.} The Influenza-like Illness (ILI) dataset captures the weekly number of reported cases involving severe influenza symptoms with complications.

\subsection {A.3: Evaluation Metrics}

We evaluate the forecasting performance using two widely adopted metrics: Mean Squared Error (MSE) and Mean Absolute Error (MAE). These metrics offer complementary perspectives on prediction accuracy. MSE penalizes larger errors more heavily due to the squaring term, making it particularly sensitive to significant deviations and thus suitable for capturing overall stability and variance in the predictions. In contrast, MAE measures the average magnitude of errors in a more uniform manner, providing a robust view of typical forecasting accuracy.

Let $N$ denote the prediction horizon, and $y_n$ and $\hat{y}_n$ represent the ground truth and predicted values at step $n$, respectively, where $n \in \{1, \dots, N\}$. The metrics are defined as follows.

\begin{equation}
\text{MSE} = \frac{1}{N} \sum_{n=1}^{N} (y_n - \hat{y}_n)^2
\end{equation}

\begin{equation}
\text{MAE} = \frac{1}{N} \sum_{n=1}^{N} |y_n - \hat{y}_n|
\end{equation}

\section {Appendix B: Implementation Details}
\subsection {B.1: Prompt Description}
\label{appendix:prompt_description}

To adapt multivariate time series data for language model processing, we design dataset-specific prompt templates that transform structured temporal data into natural language sequences. Each prompt, as illustrated in figure \ref{fig:prompt_templates}, captures a sliding window of observations and encodes four key components: the temporal interval, the sequence of numerical values, the sampling resolution, and a summary statistic representing the trend over the interval. Specifically, \textbf{[t1]} and \textbf{[t2]} indicate the start and end timestamps of the window; \textbf{[value1, ..., valuen]} denotes the ordered sequence of measurements; \textbf{[f]} specifies the data sampling frequency; and \textbf{[T]} encodes a high-level trend metric (e.g., cumulative change) over the window. This textualization enables the integration of time series dynamics into language-based architectures.

\begin{figure*}[!t]
\centering
\begin{tcolorbox}[colback=lightgray!20, colframe=gray!50, sharp corners=southwest, boxrule=0.3pt, width=\textwidth]
\begin{ttfamily}
\color{black}
\noindent
\textbf{ETTm1}: From \textcolor{red}{[t1]} to \textcolor{red}{[t2]}, the values were \textcolor{blue}{[value1, ..., valuen]} every \textcolor{magenta}{[15 minutes]}. The total trend value was \textcolor{green!50!black}{[T]}.\\

\textbf{ETTm2}: From \textcolor{red}{[t1]} to \textcolor{red}{[t2]}, the values were \textcolor{blue}{[value1, ..., valuen]} every \textcolor{magenta}{[15 minutes]}. The total trend value was \textcolor{green!50!black}{[T]}.\\

\textbf{ETTh1}: From \textcolor{red}{[t1]} to \textcolor{red}{[t2]}, the values were \textcolor{blue}{[value1, ..., valuen]} every \textcolor{magenta}{[hour]}. The total trend value was \textcolor{green!50!black}{[T]}.\\

\textbf{ETTh2}: From \textcolor{red}{[t1]} to \textcolor{red}{[t2]}, the values were \textcolor{blue}{[value1, ..., valuen]} every \textcolor{magenta}{[hour]}. The total trend value was \textcolor{green!50!black}{[T]}.\\

\textbf{ECL}: From \textcolor{red}{[t1]} to \textcolor{red}{[t2]}, the values were \textcolor{blue}{[value1, ..., valuen]} every \textcolor{magenta}{[hour]}. The total trend value was \textcolor{green!50!black}{[T]}.\\

\textbf{Weather}: From \textcolor{red}{[t1]} to \textcolor{red}{[t2]}, the values were \textcolor{blue}{[value1, ..., valuen]} every \textcolor{magenta}{[10 minutes]}. The total trend value was \textcolor{green!50!black}{[T]}.\\

\textbf{Exchange}: From \textcolor{red}{[t1]} to \textcolor{red}{[t2]}, the values were \textcolor{blue}{[value1, ..., valuen]} every \textcolor{magenta}{[day]}. The total trend value was \textcolor{green!50!black}{[T]}.\\

\textbf{ILI}: From \textcolor{red}{[t1]} to \textcolor{red}{[t2]}, the values were \textcolor{blue}{[value1, ..., valuen]} every \textcolor{magenta}{[week]}. The total trend value was \textcolor{green!50!black}{[T]}.\\

\end{ttfamily}
\end{tcolorbox}
\caption{Dataset-specific prompt templates used to convert multivariate time series windows into natural language descriptions. Each prompt encapsulates the observation window's time span, value sequence, sampling frequency, and trend summary, thereby aligning structured temporal data with language model input requirements.}
\label{fig:prompt_templates}
\end{figure*}

\subsection{B.2: Hyperparameter Sensitivity}
\label{appendix:hyperparameters}

\begin{figure*}[!h]
    \centering
    \includegraphics[width=0.9\textwidth]{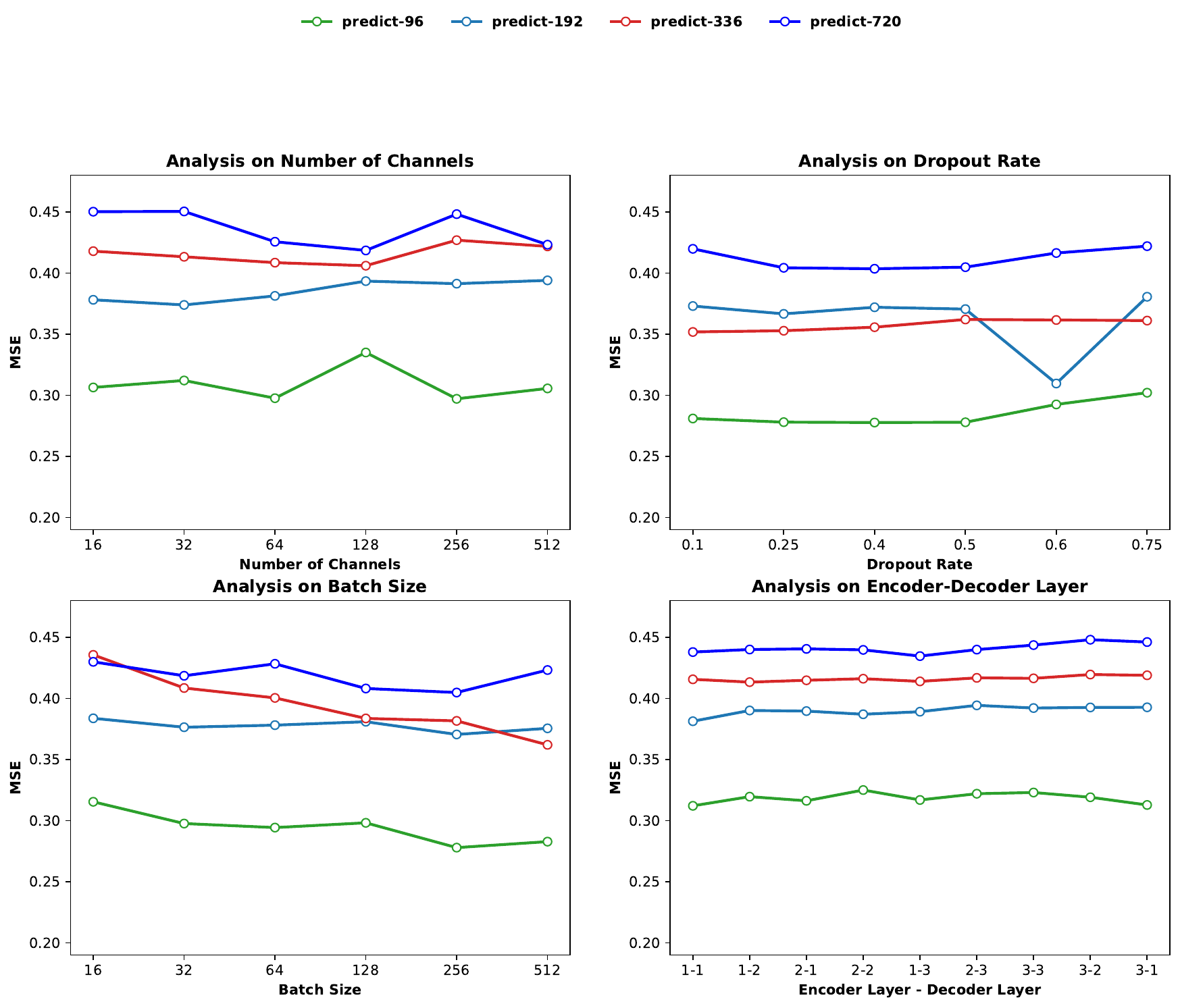}
    \caption{Sensitivity analysis of key hyperparameters on the ETTh2 dataset. We evaluate the impact of varying channel dimension, dropout rate, batch size, and encoder-decoder depth on forecasting across horizon lengths $\{96, 192, 336, 720\}$. }
    \label{fig:hyperparameters}
\end{figure*}

For all experiments, we employed grid search to identify the optimal hyperparameters of our model across different datasets. The complete hyperparameter search space is detailed in Table~\ref{tab:hyperparameter-search}. For each dataset, we selected the set of hyperparameters that yielded the lowest validation loss and reported the corresponding test performance. Optimal hyperparameters are provided in the codebase. The MSE sensitivity of different hyperparameters are also illustrated in Figure \ref{fig:hyperparameters} for the ETTh2 dataset across varying forecasting horizons, $H \in\{96, 192, 336, 720\}$.

\begin{table*}[!ht]
\centering
\begin{tabular}{l | l}
\hline
\textbf{Hyperparameter} & \textbf{Search Space} \\
\hline
Encoder Layer ($e\_layer$) & \{1, 2, 3, 4, 5, 6\} \\
Decoder Layer ($d\_layer$) & \{1, 2, 3, 4, 5, 6\} \\
Channel Dimension ($channel$) & \{16, 32, 64, 128, 256\} \\
Batch Size ($batch\_size$) & \{8, 16, 32, 64, 128, 256\} \\
Dropout Rate ($dropout$) & \{0.1, 0.2, 0.3, 0.4, 0.5, 0.6\} \\
\hline
\end{tabular}
\caption{ Hyperparameter search space used for model optimization.}
\label{tab:hyperparameter-search}
\end{table*}

\subsection {B.3: Model Configurations}
\label{model_configurations}

The architectural and training configurations adopted for our framework across diverse datasets are summarized in Tab.~\ref{tab:exp_config}.
The input length is uniformly set to $96$ time steps across all tasks, providing a consistent temporal context for forecasting. A multi-head attention mechanism with 4 heads is used consistently for all configurations, and dropout rates are tuned individually to mitigate overfitting.
Training-related parameters, presented in the five rightmost columns of Tab.~\ref{tab:exp_config}, include dataset-specific batch sizes and epoch counts tailored to convergence behaviour. The model is trained using the MSE loss across all experiments. The learning rate and weight decay are kept fixed throughout all the experiments. Individual horizon specific hyperparameters are detailed in the codebase. \\ \newline


\begin{table*}[!ht]
\centering
\scriptsize
\setlength{\tabcolsep}{4.5pt}
\renewcommand{\arraystretch}{1.2}
\begin{tabular}{c|c|c|c|c|c|c|c|c|c|c|c}
\toprule
\multirow{2}{*}{\diagbox[width=12em, height=4em]{\textbf{Dataset}}{\textbf{Configuration}}} & 
\multicolumn{6}{c|}{\textbf{Model Hyperparameter}} & 
\multicolumn{5}{c}{\textbf{Training Process}} \\
\cmidrule{2-12}
& \textbf{Encoder Layer} & \textbf{Decoder Layer} & \textbf{Input} & \textbf{Channel Dim.} & \textbf{Heads} & \textbf{Dropout} & \textbf{Learning Rate} & \textbf{Weight Decay} & \textbf{Loss} & \textbf{Batch Size} & \textbf{Epochs} \\
\midrule
ETTm1   & 1 & 2 & 96 & 128 & 4 & 0.5  & 1e-4 & 1e-3 & MSE & 64  & 150 \\
ETTm2   & 1 & 1 & 96 & 64  & 4 & 0.6  & 1e-4 & 1e-3 & MSE & 16  & 150 \\
ETTh1   & 1 & 1 & 96 & 256 & 4 & 0.4  & 1e-4 & 1e-3 & MSE & 256 & 150 \\
ETTh2   & 1 & 1 & 96 & 64  & 4 & 0.25 & 1e-4 & 1e-3 & MSE & 256 & 150 \\
ECL     & 1 & 2 & 96 & 128 & 4 & 0.3  & 1e-4 & 1e-3 & MSE & 128 & 50  \\
Weather & 6 & 2 & 96 & 64  & 4 & 0.1  & 1e-4 & 1e-3 & MSE & 32  & 150 \\
ILI     & 1 & 1 & 96 & 32  & 4 & 0.1  & 1e-4 & 1e-3 & MSE & 16  & 100 \\
\bottomrule
\end{tabular}
\caption{An overview of the experimental configurations adopted for our model.}
\label{tab:exp_config}
\end{table*}

\subsection{B.4: Multi-Head CMA}

\begin{table*}[!ht]
\centering
\scriptsize  
\setlength{\tabcolsep}{4.5pt}
\renewcommand{\arraystretch}{1.2}

\begin{tabular}{c|c|cc|cc|cc|cc|cc}
\toprule
\multirow{2}{*}{\textbf{Dataset}} & \multirow{2}{*}{\textbf{Horizon}} & 
\multicolumn{2}{c|}{\textbf{Head = 4}} & 
\multicolumn{2}{c|}{\textbf{Head = 1}} & 
\multicolumn{2}{c|}{\textbf{Head = 2}} & 
\multicolumn{2}{c|}{\textbf{Head = 8}} & 
\multicolumn{2}{c}{\textbf{Head = 16}}  \\
& & MSE & MAE & MSE & MAE & MSE & MAE & MSE & MAE & MSE & MAE \\
\midrule

\multirow{4}{*}{ETTm1} 
& 96  & \textbf{{0.308}}&0.354 & 0.311&0.357 & \textbf{{0.308}}&\textbf{{0.353}}& 0.311&0.357 & 0.313&0.359\\
& 192 & 0.357&\textbf{{0.381}} & 0.356&0.382 & 0.356&0.382 & \textbf{{0.354}}	&\textbf{{0.381}}	&0.356	&0.383\\
& 336 & 0.382&\textbf{{0.400}} & 0.385&0.402 & 0.384&0.402 & 0.383	&0.402	&\textbf{{0.381}}	&0.402\\
& 720 &  0.442&0.437 & 0.444&0.439 & 0.441&\textbf{{0.436}} & \textbf{{0.440}}	&0.437	&0.449	&0.442\\
& Avg &  \textbf{{0.372}}&\textbf{{0.393}} & 0.374&0.395 & \textbf{{0.372}}&\textbf{{0.393}} & \textbf{{0.372}}	&0.394	&0.374	&0.396\\
\midrule

\multirow{4}{*}{ETTm2} 
& 96  & \textbf{{0.172}}&\textbf{{0.254}} & 0.175&0.258	&0.175&0.256  & 0.175&0.256 & 0.174&0.257\\
& 192 & \textbf{{0.237}}&\textbf{{0.300}} & 0.243&0.301	&0.244&0.302 & 0.244&0.303 & 0.243&0.302\\
& 336 & 0.306&\textbf{{0.337}} & 0.308&0.340	&0.308&0.343 & \textbf{{0.305}}&0.342 & \textbf{{0.305}}&0.338\\
& 720 & 0.400&0.398 & 0.406&0.402	&\textbf{{0.402
}}&\textbf{{0.399}} & 0.405&0.401 & 0.407 & 0.401     \\
& Avg & \textbf{{0.279}}&\textbf{{0.322}} & 0.283&0.325	&0.282&0.325 & 0.282&0.326 & 0.282 & 0.324     \\
\midrule

\multirow{4}{*}{ETTh1} 
& 96  & \textbf{{0.371}}&\textbf{{0.397}} & \textbf{{0.371}}&0.399	& 0.374&0.399 & \textbf{{0.371}}&\textbf{{0.397}} & 0.373&0.398\\
& 192 & \textbf{{0.411}}&\textbf{{0.421}} & \textbf{{0.411}}&\textbf{{0.421}}	& \textbf{{0.411}}&\textbf{{0.421}} & 0.412&0.423 & 0.412&0.424\\
& 336 & 0.448&\textbf{{0.441}} & 0.454&0.444 & 0.451& 0.443&0.453 & 0.445& \textbf{{0.447}}	&0.444\\
& 720 & \textbf{{0.441}}&\textbf{{0.460}} & 0.447&0.463 & 0.448& 0.464&0.444 & 0.462& 0.445&0.464\\
& Avg & \textbf{{0.418}}&\textbf{{0.430}} & 0.421&0.432 & 0.421& 0.432&0.419 & 0.432& 0.419&0.433\\
\midrule

\multirow{4}{*}{ETTh2} 
& 96  & \textbf{{0.278}}&0.338 & 0.287&0.342 & 0.287&0.342 & 0.288&0.344 & 0.285&0.341 \\
& 192 & 0.351&0.389 & 0.367&0.397 & 0.367&0.398 & \textbf{{0.343}}&\textbf{{0.385}}	& 0.370&0.402 \\
& 336 & 0.358&\textbf{{0.398}} & 0.362&0.400 & 0.361&0.400 & \textbf{{0.355}}&\textbf{{0.396}}	& 0.359&0.402 \\
& 720 & 0.404&\textbf{{0.433}} & 0.402&0.431 & 0.404&0.436 & 0.402&\textbf{{0.430}} & \textbf{{0.400}}&0.433 \\
& Avg & 0.348&\textbf{{0.390}} & 0.355&0.392 & 0.355&0.394 & \textbf{{0.347}}&{0.389} & 0.354&0.394 \\
\midrule

\multirow{4}{*}{Weather} 
& 96  & 0.162&0.210 & 0.167	&0.212	&0.166	&0.212 & \textbf{{0.161}}&\textbf{{0.208}}	&0.168	&0.215\\
& 192 & \textbf{{0.211}}&\textbf{{0.253}} & 0.212	&\textbf{{0.253}}	&0.213	&0.254 &0.215	&0.258	&0.215	&0.260\\
& 336 & \textbf{{0.267}}&\textbf{{0.293}} & 0.276	&0.298	&0.269	&0.295 &0.269	&0.297	&0.272	&0.297\\
& 720 & \textbf{{0.335}}&0.346 & 0.344	&0.345	&0.339	&\textbf{{0.343}} &0.342	&0.349 & 0.344 & 0.344 \\
& Avg & \textbf{{0.244}}&\textbf{{0.275}} & 0.249	&0.277	&0.247	&0.276 &0.247	&0.278 & 0.250 & 0.279 \\
\midrule

\multirow{4}{*}{Exchange} 
& 96  &  0.085&0.205 & 0.085&0.205 & 0.085&0.205 & \textbf{{0.084}}&\textbf{{0.204}} & 0.087&0.207\\
& 192 &  0.172&0.296 & 0.174&0.298 & 0.172&0.296 & \textbf{{0.170}}&\textbf{{0.295}} & 0.175&0.298\\
& 336 &  \textbf{{0.318}}&\textbf{{0.408}} & 0.321&0.412 & 0.332&0.420 & 0.324&0.414	& 0.323&0.412\\
& 720 &  \textbf{{0.836}}&\textbf{{0.696}} & 0.933&0.737 & 0.842&0.698 & 0.876&0.712	& 0.955&0.744\\
& Avg &  \textbf{{0.353}}&\textbf{{0.401}} & 0.378&0.413 & 0.358&0.405 & 0.364&0.406	& 0.385&0.415\\
\midrule

\multirow{4}{*}{ILI} 
& 24  & 1.583&0.802 & 1.676&0.807	& 1.552&0.801 &  1.567&\textbf{{0.784}}	& \textbf{{1.542}}&0.791\\
& 36  & 1.601&0.820 & \textbf{{1.510}}&0.806	& 1.569&\textbf{{0.802}} & 1.594&0.812	& 1.734&0.852\\
& 48  & \textbf{{1.718}}&\textbf{{0.815}} & 2.114&0.935	& 1.774&0.849 &  1.826&0.873	& 1.827&0.874\\
& 60  & \textbf{{1.920}}&\textbf{{0.901}} & 1.955&0.912	& 2.003&0.922 &  1.973&0.912	& 1.984&0.911\\
& Avg & \textbf{{1.705}}&\textbf{{0.835}} & 1.814&0.865	& 1.725&0.844 & 1.739&0.845	& 1.772&0.857\\
\midrule

\multicolumn{2}{c|}{\textbf{1st Count}} & 20&24 & 3&2  &  4&7  &  11&10  &  5&1     \\

\bottomrule
\end{tabular}
\caption{Ablation study on the number of heads in the Cross-Modal Attention (CMA) module across benchmark datasets and prediction lengths. Metrics reported are MSE and MAE (lower is better). Input sequence length is set to $96$. \textbf{Bold:} the best.}
\label{tab:multi-head cma}
\end{table*}

To investigate the impact of the number of heads in the Multi-Head Cross-Modal Attention (CMA) module, we evaluated five different configurations: $\{1, 2, 4, 8, 16\}$, keeping all other architectural components constant. As reported in Table~\ref{tab:multi-head cma}, the 4-head configuration consistently delivers the best performance, achieving the lowest Mean Squared Error (MSE) and Mean Absolute Error (MAE) in 21 and 20 out of 35 benchmark cases, respectively. 

The 8-head variant shows relatively competitive performance, securing the best score in 10 out of 35 cases for both MSE and MAE. However, increasing the head count beyond this (e.g., to 16) does not lead to further improvements and often degrades performance—likely due to over-fragmentation of attention or redundancy in representation subspaces. On the other hand, the 1-head and 2-head settings underperform across most datasets and horizons, indicating insufficient capacity for capturing complex cross-modal alignments. Overall, the 4-head CMA setting achieves the lowest average MSE and lowest average MAE across all tasks, validating its effectiveness in balancing representation richness and attention diversity. Consequently, we adopt 4 heads as the default configuration for the CMA module in T3Time.


\begin{table*}[!ht]
\centering
\scriptsize 
\setlength{\tabcolsep}{4.5pt}
\renewcommand{\arraystretch}{0.8}

\begin{tabular}{c|c|cc|cc|cc|cc|cc|cc|cc|cc|cc}
\toprule
\multirow{2}{*}{\textbf{Dataset}} & \multirow{2}{*}{\textbf{Horizon}} & 
\multicolumn{2}{c|}{\textbf{Ours}} & 
\multicolumn{2}{c|}{\textbf{TimeCMA}} & 
\multicolumn{2}{c|}{\textbf{TimeLLM}} & 
\multicolumn{2}{c|}{\textbf{UniTime}} & 
\multicolumn{2}{c|}{\textbf{TimesNet}} & 
\multicolumn{2}{c|}{\textbf{DLinear}} & 
\multicolumn{2}{c|}{\textbf{iTransformer}} & 
\multicolumn{2}{c|}{\textbf{PatchTST}} & 
\multicolumn{2}{c}{\textbf{OFA}} \\
& & MSE & MAE & MSE & MAE & MSE & MAE & MSE & MAE & MSE & MAE & MSE & MAE & MSE & MAE & MSE & MAE& MSE & MAE \\
\midrule

\multirow{4}{*}{ETTm1} 
& 96  &  \textbf{{0.308}}&\underline{0.354}  &  \underline{0.312}&\textbf{{0.351}} & 0.359&0.381 & 0.322&0.363 & 0.338&0.375 & 0.345&0.372 & 0.334&0.368 & 0.344&0.373 & 0.335&0.369 \\
& 192 &  \textbf{{0.357}}&\underline{0.381}  &  \underline{0.361}&\textbf{{0.378}} & 0.383&0.393 & 0.366&0.387 & 0.374&0.387 & 0.380&0.389 & 0.377&0.391 & 0.367&0.386 & 0.374&0.385 \\
& 336 &  \textbf{{0.382}}&\textbf{{0.400}}  &  \underline{0.392}&\underline{0.401} & 0.416&0.414 & 0.398&0.407 & 0.410&0.411 & 0.413&0.413 & 0.426&0.420 & 0.392&0.407 & 0.407&0.406 \\
& 720 &  \textbf{{0.442}}&\textbf{{0.437}}  &  \underline{0.453}&\underline{0.438} & 0.483&0.449 & 0.454&0.440 & 0.478&0.450 & 0.474&0.453 & 0.491&0.459 & 0.464&0.442 & 0.469&0.442 \\
& Avg &  \textbf{{0.372}}&\underline{0.393}  &  \underline{0.380}&\textbf{{0.392}} & 0.410&0.409 & 0.385&0.399 & 0.400&0.406 & 0.403&0.407 & 0.407&0.410 & 0.392&0.402 & 0.396&0.401 \\
\midrule

\multirow{4}{*}{ETTm2} 
& 96  &  \textbf{{0.172}}&\textbf{{0.254}}  &  \underline{0.173}&\underline{0.258} &  0.193&0.280 & 0.183&0.266 &  0.187&0.267 & 0.193&0.292 & 0.180&0.264 & 0.177&0.260 & 0.190&0.275 \\
& 192 &  \textbf{{0.237}}&\textbf{{0.300}}  &  \underline{0.238}&\underline{0.301} &  0.257&0.318 & 0.251&0.310  &  0.249&0.309 & 0.284&0.362 & 0.250&0.309 & 0.246&0.305 & 0.253&0.313 \\
& 336 &  0.306& \textbf{{0.337}}  &  \textbf{{0.297}}&\underline{0.338} &  0.317&0.353 & 0.183&0.266 &  0.319&0.351 & 0.369&0.427 & 0.311&0.348 & \underline{0.305}&0.343 & 0.321&0.360 \\
& 720 &  \underline{0.400}&\underline{0.398}  &  \textbf{{0.393}}&\textbf{{0.394}} &  0.419&0.411 & 0.420&0.410 &  0.408&0.403 & 0.554&0.522 & 0.412&0.407 & 0.410&0.405 & 0.411&0.406\\
& Avg &  \underline{0.279}&\textbf{{0.322}}  &  \textbf{{0.275}}&\textbf{{0.323}} &  0.296&0.340 & 0.293&0.334 &  0.291&0.333 & 0.350&0.401 & 0.288&0.332 & 0.285&\underline{0.328} & 0.294&0.339\\
\midrule

\multirow{4}{*}{ETTh1} 
& 96 &   \textbf{{0.371}}&\underline{0.397}  &  \underline{0.373}&\textbf{{0.391}} & 0.398&0.410 & 0.397&0.418 & 0.384&0.402 & 0.386&0.400 & 0.386&0.405 & 0.404&0.413 & 0.398&0.424 \\
& 192 &  \textbf{{0.411}}& \textbf{{0.421}}  &  \underline{0.427}&\textbf{{0.421}} & 0.451&0.440 & 0.434&0.439 & 0.434& \underline{0.429} & 0.437&0.432 & 0.441&0.436 & 0.454&0.430 & 0.449&0.427 \\
& 336 &  \textbf{{0.448}}&\textbf{{0.441}}  &  \underline{0.458}&\underline{0.448} & 0.473&0.451 & 0.468&0.457 & 0.491&0.469 & 0.481&0.459 & 0.487&0.458 & 0.497&0.462 & 0.492&0.466 \\
& 720 &  \textbf{{0.441}}&\textbf{{0.460}}  &  \underline{0.449}&\textbf{{0.460}} & 0.469& \underline{0.470} & 0.469&0.477 & 0.521&0.500 & 0.519&0.516 & 0.503&0.491 & 0.496&0.481 & 0.487&0.483 \\
& Avg &  \textbf{{0.418}}&\textbf{{0.430}}  &  \underline{0.423}&\underline{0.431} & 0.448&0.443 & 0.442&0.448 & 0.458&0.450 & 0.456&0.452 & 0.454&0.447 & 0.463&0.449 & 0.457&0.450 \\
\midrule

\multirow{4}{*}{ETTh2} 
& 96  & \textbf{{0.278}}&\underline{0.338} & \underline{0.286}&\textbf{{0.336}} & 0.295&0.345 & 0.296&0.345 & 0.340&0.374 & 0.333&0.387 & 0.297&0.349 & 0.312&0.358 & 0.312&0.360 \\
& 192 & \textbf{{0.351}}&\underline{0.389} & \underline{0.363}&\textbf{{0.387}} & 0.386&0.399 & 0.374&0.394 & 0.402&0.414 & 0.477&0.476 & 0.380&0.400 & 0.397&0.408 & 0.387&0.405 \\
& 336 & \textbf{{0.358}}&\textbf{{0.398}} & \underline{0.406}&\underline{0.421} & 0.419&0.429 & 0.415&0.427 & 0.452&0.452 & 0.594&0.541 & 0.428&0.432 & 0.435&0.440 & 0.424&0.437 \\
& 720 & \textbf{{0.404}}&\textbf{{0.433}} & \underline{0.417}&\underline{0.438} & 0.425&0.442 & 0.425&0.444 & 0.462&0.468 & 0.831&0.657 & 0.427&0.445 & 0.436&0.449 & 0.433&0.453 \\
& Avg & \textbf{{0.348}}&\textbf{{0.390}} & \underline{0.372}&\underline{0.397} & 0.381&0.404 & 0.378&0.403 & 0.414&0.427 & 0.559&0.515 & 0.383&0.407 & 0.395&0.414 & 0.389&0.414 \\
\midrule

\multirow{4}{*}{ECL}  
& 96  & \textbf{{0.138}}&\textbf{{0.233}} & \underline{0.143}&\underline{0.238} & 0.172&0.265 & 0.196&0.287 &  0.168&0.272 & 0.197&0.282 & 0.148&0.240 & 0.186&0.269 & 0.197&0.290 \\
& 192 & \textbf{{0.155}}&\textbf{{0.250}} & \underline{0.161}&0.259 & 0.182&0.279 & 0.199&0.291 &  0.184&0.289 & 0.196&0.285 & 0.162&\underline{0.253} & 0.190&0.273 & 0.201&0.292 \\
& 336 & \textbf{{0.168}}&\underline{0.265} & 0.169&\textbf{{0.261}} & 0.195&0.288 & 0.214&0.305 &  0.198&0.300 & 0.209&0.301 & 0.178&0.269 & 0.206&0.290 & 0.217&0.309 \\
& 720 & \textbf{{0.218}}&\textbf{{0.314}} & \underline{0.219}&\underline{0.315} & 0.233&0.320 & 0.254&0.335 &  0.220&0.320 & 0.245&0.333 & 0.225&0.317 & 0.247&0.322 & 0.253&0.339 \\
& Avg & \textbf{{0.170}}&\textbf{{0.266}} & 0.174&0.269 & 0.195&0.288 & 0.216&0.306 &  0.192&0.295 & 0.212&0.300 & 0.178&0.270 & 0.207&0.289 & 0.217&0.308\\
\midrule

\multirow{4}{*}{Weather} 
& 96  & \textbf{{0.162}}&\textbf{{0.210}} & \underline{0.167}&\underline{0.211} & 0.198&0.235 & 0.171&0.214 & 0.172&0.220 & 0.196&0.255 & 0.174&0.214 & 0.177&0.218 & 0.203& 0.244 \\
& 192 & \textbf{{0.211}}&\textbf{{0.253}} & \textbf{{0.212}}&\textbf{{0.253}} & 0.240&0.269 & \underline{0.217}& \underline{0.254} & 0.219&0.261 & 0.237&0.296 & 0.221&0.254 & 0.222&0.259 & 0.247&0.277 \\
& 336 & \textbf{{0.267}}&\underline{0.293} & \underline{0.270}&\textbf{{0.292}} & 0.295&0.308 & 0.274&0.293 & 0.280&0.306 & 0.283&0.335 & 0.278&0.296 & 0.277&0.297 & 0.297&0.311 \\
& 720 & \textbf{{0.335}}&\textbf{{0.346}} & 0.350&\underline{0.348} & 0.368&0.353 & 0.351&0.343 & 0.365&0.359 & \underline{0.345}&0.381 & 0.358&0.349 & 0.352&0.347 & 0.368&0.356 \\
& Avg & \textbf{{0.244}}&\textbf{{0.275}} & \underline{0.250}&\underline{0.276} & 0.275&0.291 & 0.253&\underline{0.276} & 0.259&0.287 & 0.265&0.317 & 0.258&0.278 & 0.257&0.280 & 0.279&0.297 \\
\midrule

\multirow{4}{*}{ILI} 
& 24  & \textbf{{1.583}}&\textbf{{0.802}} & \underline{1.996}&0.998 & 2.383&1.004 & 2.346&0.954 &  2.317&\underline{0.934} & 2.398&1.040 & 2.347&1.731 & 2.335&0.989 & 2.732&1.100 \\
& 36  & \textbf{{1.601}}&\textbf{{0.820}} & \underline{1.906}&0.915 & 2.390&0.993 & 1.998&0.912 &  1.972&0.920 & 2.646&1.088 & 2.468&0.998 & 2.561&1.035 & 2.664&1.063 \\
& 48  & \textbf{{1.718}}&\textbf{{0.815}} & \underline{1.867}&\underline{0.868} & 2.394&1.003 & 1.979&0.912 &  2.238&0.913 & 2.614&1.086 & 2.489&1.016 & 2.465&1.022 & 2.617&1.041 \\
& 60  & \textbf{{1.920}}&\textbf{{0.901}} & \textbf{{1.920}}&\underline{0.904} & 2.562&1.049 & \underline{2.109}&0.938 &  2.027&0.928 & 2.804&1.146 & 2.471&1.065 & 2.189&0.997 & 2.478&1.035 \\
& Avg & \textbf{{1.705}}&\textbf{{0.835}} & \underline{1.922}&\underline{0.921} & 2.432&1.012 & 2.108&0.929 &  2.139&0.931 & 2.616&1.090 & 2.444&1.203 & 2.388&1.011 & 2.623&1.060 \\
\midrule

\multirow{4}{*}{Exchange} 
& 96  & \textbf{{0.085}}&\textbf{{0.205}} & 0.099 & 0.224 & 0.087 & 0.208 & \underline{0.086} & 0.209 &  0.107 & 0.234 & 0.088 & 0.218 & \underline{0.086} & \underline{0.206} & 0.109 & 0.236 & 0.148 & 0.278 \\
& 192  & \textbf{{0.172}}&\textbf{{0.296}} & 0.186& 0.312 & 0.173 & 0.299 & \underline{0.174} & \underline{0.299}  & 0.226 & 0.344 & 0.176 & 0.315 & 0.177& \underline{0.299} & 0.205 & 0.327 & 0.271 & 0.380 \\
& 336  & \underline{0.318}&\textbf{{0.408}} & 0.364& 0.444 & 0.375 & 0.454 & 0.319 & \textbf{{0.408}} &  0.367 & 0.448 & \textbf{{0.313}} & 0.427 & 0.331 & 0.417 & 0.356 & 0.436 & 0.460 & 0.500 \\
& 720  & \textbf{{0.836}} & 0.696 & 0.932 & 0.735 & 0.853 & 0.703 & 0.875 & 0.701  &  0.964 & 0.746 & \underline{0.839} & \underline{0.695} & 0.847 & \textbf{{0.691}} & 0.888 & 0.716 & 1.195 & 0.841 \\
& Avg & \textbf{{0.353}}&\textbf{{0.401}} & 0.395 & 0.429 & 0.372 & 0.416 & 0.364 & 0.404 &  0.416& 0.443 & \underline{0.354} & 0.414 & 0.360 & \underline{0.403} & 0.390 & 0.429 & 0.519 & 0.500 \\
\midrule

\multicolumn{2}{c|}{\textbf{1st Count}} &  \textbf{{36}}&\textbf{{30}}  &  \underline{5}&\underline{12} &  0&0  &  0&1  &  0&0  &  1&0  &  0&1  &  0&0  &  0&0  \\

\bottomrule
\end{tabular}

\caption{Multivariate long-term forecasting results. Input sequence is set to $96$ for all but ILI (input sequence $36$). The prediction horizon for all bechmark datasets, $H \in \{96, 192, 336, 720\}$, and for ILI, $H \in \{24, 36, 48, 60\}$. A lower score signals better performance. \textbf{Bold}: the best, \underline{underline}: the second best.}
\label{tab:multivariate_results}
\end{table*}

\section{Appendix C: Long Term Forecasting Details}
The complete results for the long term forecasting tasks of our model are presented in Table \ref{tab:multivariate_results}. For all benchmark datasets, the input sequence length is fixed at $96$, and the prediction sequence lengths are set to $\{96, 192, 336, 720\}$. However, for the ILI dataset, the input sequence is set to $36$ and the output sequence length is $\{24, 36, 48, 60\}$. In the long-term forecasting scenario, our model achieves state-of-the-art (SOTA) performance in 66 out of 80 cases across 8 diverse time series benchmarks.

\section{Appendix D: Few Shot Forecasting Details}
The complete results for the few-shot forecasting tasks of our model are presented in Tables \ref{tab:few_shot_10_full} and \ref{tab:few_shot_5_full}. For all benchmark datasets, the input sequence length is fixed at $512$, and the prediction sequence lengths are set to $\{96, 192, 336, 720\}$. In the 10\% few-shot scenario, our model achieves state-of-the-art (SOTA) performance in 43 out of 50 cases across five diverse time series benchmarks. Furthermore, in the 5\% few-shot setting, our model attains SOTA performance in 38 out of 46 cases across the same five benchmarks.

\begin{table*}[!ht]
\centering
\scriptsize  
\setlength{\tabcolsep}{4.5pt}
\renewcommand{\arraystretch}{0.8}

\begin{tabular}{c|c|cc|cc|cc|cc|cc|cc|cc|cc}
\toprule
\multirow{2}{*}{\textbf{Dataset}} & \multirow{2}{*}{\textbf{Horizon}} & 
\multicolumn{2}{c|}{\textbf{Ours}} & 
\multicolumn{2}{c|}{\textbf{TimeCMA}} & 
\multicolumn{2}{c|}{\textbf{TimeLLM}} & 
\multicolumn{2}{c|}{\textbf{GPT4TS}} & 
\multicolumn{2}{c|}{\textbf{TimesNet}} & 
\multicolumn{2}{c|}{\textbf{DLinear}} & 
\multicolumn{2}{c|}{\textbf{PatchTST}} & 
\multicolumn{2}{c}{\textbf{Fedformer}} \\
& & MSE & MAE & MSE & MAE & MSE & MAE & MSE & MAE & MSE & MAE & MSE & MAE & MSE & MAE & MSE & MAE\\
\midrule

\multirow{4}{*}{ETTm1} 
& 96  & \textbf{0.315}&\textbf{0.359} & \underline{0.339}&\underline{0.380} & 0.346&0.388 & 0.390&0.404 & 0.583&0.501 & 0.352&0.392 & 0.410&0.419 & 0.578&0.518\\
& 192 & \textbf{0.350}&\textbf{0.383} & \underline{0.373}&\underline{0.404} & \underline{0.373}&0.416 & 0.429&0.423 & 0.630&0.528 & 0.382&0.412 & 0.437&0.434 & 0.617&0.546\\
& 336 & \textbf{0.390}&\textbf{0.406} & \textbf{0.390}&\underline{0.412} & \underline{0.413}&0.426 & 0.469&0.439 & 0.725&0.568 & 0.419&0.434 & 0.476&0.454 & 0.998&0.775\\
& 720 & \textbf{0.447}&\textbf{0.442} & \underline{0.448}&\underline{0.444} & 0.485&0.476 & 0.569&0.498 & 0.769&0.549 & 0.490&0.477 & 0.681&0.556 & 0.693&0.579\\
& Avg & \textbf{0.376}&\textbf{0.398} & \underline{0.387}&\underline{0.410} & 0.404&0.427 & 0.464&0.441 & 0.677&0.537 & 0.411&0.429 & 0.501&0.466 & 0.722&0.605\\
\midrule

\multirow{4}{*}{ETTm2} 
& 96  & \textbf{0.176}&\underline{0.268} & 0.219&0.303 & \underline{0.177}&\textbf{0.261} & 0.188&0.269 & 0.212&0.285 & 0.213&0.303 & 0.191&0.274 & 0.291&0.399\\
& 192 & \textbf{0.233}&\textbf{0.307} & 0.272&0.333 & \underline{0.241}&\underline{0.314} & 0.251&0.309 & 0.270&0.323 & 0.278&0.345 & 0.252&0.317 & 0.307&0.379\\
& 336 & \underline{0.283}&\underline{0.341} & 0.340&0.375 & \textbf{0.274}&\textbf{0.327} & 0.307&0.346 & 0.323&0.353 & 0.338&0.385 & 0.306&0.353 & 0.543&0.559\\
& 720 & \textbf{0.373}&\underline{0.394} & 0.418&0.422 & \underline{0.417}&\textbf{0.390} & 0.426&0.417 & 0.474&0.449 & 0.436&0.440 & 0.433&0.427 & 0.712&0.614\\
& Avg & \textbf{0.266}&\underline{0.327} & 0.312&0.358 & \underline{0.277}&\textbf{0.323} & 0.293&0.335 & 0.320&0.353 & 0.316&0.368 & 0.296&0.343 & 0.463&0.488\\
\midrule

\multirow{4}{*}{ETTh1} 
& 96  & \textbf{0.412}&\textbf{0.422} & 0.453&0.459 & \underline{0.448}&0.460 & 0.458&\underline{0.456} & 0.861&0.628 & 0.492&0.495 & 0.516&0.485 & 0.512&0.499\\
& 192 & \textbf{0.440}&\textbf{0.450} & \underline{0.459}&\underline{0.464} & 0.484&0.483 & 0.570&0.516 & 0.797&0.593 & 0.565&0.538 & 0.598&0.524 & 0.624&0.555\\
& 336 & \textbf{0.463}&\textbf{0.465} & \underline{0.484}&\underline{0.484} & 0.589&0.540 & 0.608&0.535 & 0.941&0.648 & 0.721&0.622 & 0.657&0.550 & 0.691&0.574\\
& 720 & \textbf{0.481}&\textbf{0.481} & \underline{0.526}&\underline{0.508} & 0.700&0.604 & 0.725&0.591 & 0.877&0.641 & 0.986&0.743 & 0.762&0.610 & 0.728&0.614\\
& Avg & \textbf{0.449}&\textbf{0.454} & \underline{0.480}&\underline{0.479} & 0.556&0.522 & 0.590&0.525 & 0.869&0.628 & 0.691&0.600 & 0.633&0.542 & 0.639&0.561\\
\midrule

\multirow{4}{*}{ETTh2} 
& 96  & \underline{0.300}&\textbf{0.311} & 0.365&0.406 & \textbf{0.275}&\underline{0.326} & 0.331&0.374 & 0.378&0.409 & 0.357&0.411 & 0.353&0.389 & 0.382&0.416\\
& 192 & \textbf{0.364}&\underline{0.388} & 0.424&0.443 & \underline{0.374}&\textbf{0.373} & 0.402&0.411 & 0.490&0.467 & 0.569&0.519 & 0.403&0.414 & 0.478&0.474\\
& 336 & \textbf{0.352}&\textbf{0.407} & 0.385&\underline{0.429} & 0.406&\underline{0.429} & 0.406&0.433 & 0.537&0.494 & 0.671&0.572 & 0.426&0.441 & 0.504&0.501\\
& 720 & \textbf{0.413}&\textbf{0.446} & \underline{0.418}&0.451 & 0.427&\underline{0.449} & 0.449&0.464 & 0.510&0.491 & 0.824&0.648 & 0.477&0.480 & 0.499&0.509\\
& Avg & \textbf{0.357}&\textbf{0.388} & 0.398&0.433 & \underline{0.370}&\underline{0.394} & 0.397&0.421 & 0.479&0.465 & 0.605&0.538 & 0.415&0.431 & 0.466&0.475\\
\midrule

\multirow{4}{*}{Weather} 
& 96  & \textbf{0.154}&\textbf{0.210} & \underline{0.157}& \underline{0.214} & 0.161&{0.210} & 0.163&0.215 & 0.184&0.230 & 0.171&0.224 & 0.165&0.215 & 0.188&0.253\\
& 192 & \textbf{0.195}&\textbf{0.248} & \underline{0.198}&{0.248} & 0.204&{0.248} & 0.210&\underline{0.254} & 0.245&0.283 & 0.215&0.263 & 0.210&0.257 & 0.250&0.304\\
& 336 & \textbf{0.246}&\textbf{0.286} & \underline{0.247}&\underline{0.287} & 0.261&0.302 & 0.256&0.292 & 0.305&0.321 & 0.258&0.299 & 0.259&0.297 & 0.312&0.346\\
& 720 & \textbf{0.309}&\textbf{0.331} & \underline{0.318}&0.337 & {0.309}&\underline{0.332} & 0.321&0.339 & 0.381&0.371 & 0.320&0.346 & 0.332&0.346 & 0.387&0.393\\
& Avg & \textbf{0.226}&\textbf{0.268} & \underline{0.229}&\underline{0.272} & 0.234&0.273 & 0.238&0.275 & 0.279&0.301 & 0.241&0.283 & 0.242&0.279 & 0.284&0.324\\
\midrule

\multicolumn{2}{c|}{\textbf{1st Count}} &  \textbf{23}&\textbf{20} &  1 & 2  &  \underline{3}&\underline{7}  &  0&0  &  0&0  &  0&0  &  0&0  &  0&0   \\

\bottomrule
\end{tabular}
\caption{Few-shot forecasting results on $10\%$ training data. Input sequence is set to $512$ for all benchmark datasets. The prediction horizon, $H \in \{96, 192, 336, 720\}$. A lower score signals better performance. \textbf{Bold}: the best, \underline{underline}: the second best.}
\label{tab:few_shot_10_full}
\end{table*}

\begin{table*}[!ht]
\centering
\scriptsize  
\setlength{\tabcolsep}{4.5pt}
\renewcommand{\arraystretch}{0.8}

\begin{tabular}{c|c|cc|cc|cc|cc|cc|cc|cc|cc}
\toprule
\multirow{2}{*}{\textbf{Dataset}} & \multirow{2}{*}{\textbf{Horizon}} & 
\multicolumn{2}{c|}{\textbf{Ours}} & 
\multicolumn{2}{c|}{\textbf{TimeCMA}} & 
\multicolumn{2}{c|}{\textbf{TimeLLM}} & 
\multicolumn{2}{c|}{\textbf{GPT4TS}} & 
\multicolumn{2}{c|}{\textbf{TimesNet}} & 
\multicolumn{2}{c|}{\textbf{DLinear}} & 
\multicolumn{2}{c|}{\textbf{PatchTST}} & 
\multicolumn{2}{c}{\textbf{Fedformer}} \\
& & MSE & MAE & MSE & MAE & MSE & MAE & MSE & MAE & MSE & MAE & MSE & MAE & MSE & MAE & MSE & MAE\\
\midrule

\multirow{4}{*}{ETTm1} 
& 96  & \underline{0.326}&\textbf{0.367} & 0.340&0.385 & \textbf{0.316}&0.377 & 0.386&0.405 & 0.606&0.518 & 0.332&0.374 & 0.399&0.414 & 0.628&0.544\\
& 192 & \textbf{0.364}&\underline{0.396} & \underline{0.372}&0.403 & 0.450&0.464 & 0.440&0.438 & 0.681&0.539 & 0.358&\textbf{0.390} & 0.441&0.436 & 0.666&0.566\\
& 336 & \textbf{0.392}&\textbf{0.411} & \underline{0.402}&0.418 & 0.450&0.424 & 0.485&0.459 & 0.786&0.597 & \underline{0.402}&\underline{0.416} & 0.499&0.467 & 0.807&0.628\\
& 720 & \textbf{0.455}&\textbf{0.446} & \underline{0.472}&\underline{0.458} & 0.483&0.471 & 0.577&0.499 & 0.796&0.593 & 0.511&0.489 & 0.767&0.587 & 0.822&0.633\\
& Avg & \textbf{0.384}&\textbf{0.405} & \underline{0.396}&\underline{0.416} & 0.425&0.434 & 0.472&0.450 & 0.717&0.561 & 0.400&0.417 & 0.526&0.476 & 0.730&0.592\\
\midrule

\multirow{4}{*}{ETTm2} 
& 96  & \underline{0.176}&\underline{0.268} & 0.218&0.303 & \textbf{0.174}&\textbf{0.261} & 0.199&0.280 & 0.220&0.299 & 0.236&0.326 & 0.206&0.288 & 0.229&0.320\\
& 192 & \underline{0.231}&\underline{0.305} & 0.298&0.347 & \textbf{0.215}&\textbf{0.287} & 0.256&0.316 & 0.311&0.361 & 0.306&0.373 & 0.264&0.324 & 0.394&0.361\\
& 336 & \underline{0.286}&\underline{0.343} & 0.386&0.399 & \textbf{0.273}&\textbf{0.330} & 0.318&0.353 & 0.338&0.366 & 0.380&0.423 & 0.334&0.367 & 0.378&0.427\\
& 720 & \textbf{0.375}&\textbf{0.405} & \underline{0.416}&0.419 & 0.433&\underline{0.412} & 0.460&0.436 & 0.509&0.465 & 0.674&0.583 & 0.454&0.432 & 0.523&0.510\\
& Avg & \textbf{0.267}&\textbf{0.330} & 0.329&0.367 & \underline{0.274}&\underline{0.323} & 0.308&0.346 & 0.344&0.372 & 0.399&0.426 & 0.314&0.352 & 0.381&0.404\\
\midrule

\multirow{4}{*}{ETTh1} 
& 96  & \textbf{0.417}&\textbf{0.435} & \underline{0.457}&\underline{0.460} & 0.483&0.464 & 0.543&0.506 & 0.892&0.625 & 0.547&0.503 & 0.557&0.519 & 0.593&0.529\\
& 192 & \textbf{0.441}&\textbf{0.447} & \underline{0.460}&\underline{0.461} & 0.629&0.540 & 0.748&0.580 & 0.940&0.665 & 0.720&0.604 & 0.711&0.570 & 0.652&0.563\\
& 336 & \textbf{0.467}&\textbf{0.470} &\underline{0.500}& 0.489 & 0.768&0.626 & 0.754&0.595 & 0.945&0.653 & 0.984&0.727 & 0.816&0.619 & 0.731&0.594\\
& 720 &  -	& -     & -& -& -&-&-&-                 & - &-        & - & - & - & -             & - &-       \\
& Avg & \textbf{0.442}&\textbf{0.451} & \underline{0.472}&\underline{0.470} & 0.627&0.543 & 0.681&0.560 & 0.925&0.647 & 0.750&0.611 & 0.694&0.569 & 0.658&0.562\\
\midrule

\multirow{4}{*}{ETTh2} 
& 96  & \textbf{0.306}&\textbf{0.367} & 0.363&0.409 & \underline{0.336}&\underline{0.397} & 0.376&0.421 & 0.409&0.420 & 0.442&0.456 & 0.401&0.421 & 0.390&0.424\\
& 192 & \textbf{0.369}&\textbf{0.409} & \underline{0.404}&0.434 & 0.406&0.425 & 0.418&0.441 & 0.483&0.464 & 0.617&0.542 & 0.452&0.455 & 0.457&0.465\\
& 336 & \textbf{0.395}&\textbf{0.432} & 0.418&0.447 & \underline{0.405}&{0.432} & 0.408& \underline{0.439} & 0.499&0.479 & 1.424&0.849 & 0.464&0.469 & 0.477&0.483\\
& 720 & -    & -    & -&- & -&-&-&-                 & - & -       & - &-& -& -                & - & -      \\
& Avg & \textbf{0.357}&\textbf{0.403} & 0.395&0.430 & \underline{0.382}&\underline{0.418} & 0.400&0.433 & 0.439&0.448 & 0.694&0.577 & 0.827&0.615 & 0.463&0.454\\
\midrule

\multirow{4}{*}{Weather} 
& 96  &  \textbf{0.154}&\textbf{0.211} & \underline{0.160}&\underline{0.216} & 0.172&0.263 & 0.175&0.230 & 0.207&0.253 & 0.184&0.242 & 0.171&0.224 & 0.229&0.309\\
& 192 &  \textbf{0.194}&\textbf{0.245} & \underline{0.199}&\underline{0.251} & 0.224&0.271 & 0.227&0.276 & 0.272&0.307 & 0.228&0.283 & 0.230&0.277 & 0.265&0.317\\
& 336 &  \textbf{0.244}&\textbf{0.286} & \underline{0.247}&\underline{0.287} & 0.282&0.321 & 0.286&0.322 & 0.313&0.328 & 0.279&0.322 & 0.294&0.326 & 0.353&0.392\\
& 720 &  \textbf{0.310}&\textbf{0.333} & \underline{0.318}&\underline{0.336} & 0.366&0.381 & 0.366&0.379 & 0.400&0.385 & 0.364&0.388 & 0.384&0.387 & 0.391&0.394\\
& Avg &  \textbf{0.226}&\textbf{0.269} & \underline{0.231}&\underline{0.273} & 0.260&0.309 & 0.263&0.301 & 0.298&0.318 & 0.263&0.308 & 0.269&0.303 & 0.309&0.353\\
\midrule

\multicolumn{2}{c|}{\textbf{1st Count}} &  \textbf{19}&\textbf{19}  &  0&1  &  \underline{4}&\underline{4} &  0&0  &  0&0  &  0&1  &  0&0 & 0&0  \\

\bottomrule
\end{tabular}
\caption{Few-shot forecasting results on $5\%$ training data. Input sequence is set to $512$ for all benchmark datasets. The prediction horizon, $H \in \{96, 192, 336, 720\}$. ‘–’ indicates that 5\% of the time series data was insufficient to form a viable training set. A lower score signals better performance. \textbf{Bold}: the best, \underline{underline}: the second best.}
\label{tab:few_shot_5_full}
\end{table*}


\begin{table*}[!t]
\centering
\scriptsize  
\setlength{\tabcolsep}{4.5pt}
\renewcommand{\arraystretch}{1.2}

\begin{tabular}{c|c|cc|cc|cc|cc|cc}
\toprule
\multirow{2}{*}{\textbf{Dataset}} & \multirow{2}{*}{\textbf{Horizon}} & 
\multicolumn{2}{c|}{\textbf{Ours}} & 
\multicolumn{2}{c|}{\textbf{w/o Frequency Module}} & 
\multicolumn{2}{c|}{\textbf{w/o Multihead CMA}} & 
\multicolumn{2}{c|}{\textbf{w/o Residual Connection}} & 
\multicolumn{2}{c}{\textbf{w/o Gating Mechanism}}  \\
& & MSE & MAE & MSE & MAE & MSE & MAE & MSE & MAE & MSE & MAE \\
\midrule

\multirow{4}{*}{ETTm1} 
& 96  & \textbf{{0.308}}&	0.354	& 0.310&\textbf{{0.353}} & 0.311&0.357 & 0.333&0.372 & 0.315&0.359\\
& 192 & 0.357&	0.381	& 0.359&\textbf{{0.380}} &	\textbf{{0.356}}&0.382	& 0.393&0.407 &	0.358&0.381\\
& 336 & \textbf{{0.382}}&	0.400	& 0.392&0.402 &	0.385&0.402	& 0.420&0.423 &	0.383&\textbf{{0.399}}\\
& 720 &  0.442&	0.437	& 0.460&0.441 &	0.444&0.439	& 0.468&0.449 &	\textbf{{0.435}}&\textbf{{0.435}}\\
& Avg & \textbf{{0.372}}&	\textbf{{0.393}}	& 0.381&0.394 &	0.374&0.395	& 0.404&0.413 &	0.373&0.394\\
\midrule

\multirow{4}{*}{ETTm2} 
& 96  & 0.172&	0.254	&0.174	&0.257	&0.175	&0.258	&0.179	&0.260	&\textbf{{0.170}}	&\textbf{{0.252}}\\
& 192 & \textbf{{0.237}}&	\textbf{{0.300}}	&0.241	&0.302	&0.243	&0.301	&0.257	&0.308	&0.245	&0.301\\
& 336 & 0.306	&\textbf{{0.337}}	&\textbf{{0.300}}	&0.338	&0.308	&0.340	&0.314	&0.347	&0.301	&0.337\\
& 720 & 0.400	& 0.398	&\textbf{{0.400}}	&\textbf{{0.398}}	&0.406	&0.402	&0.402	&0.402	&0.403	&0.399\\
& Avg & \textbf{{0.279}}&	\textbf{{0.322}}	& \textbf{{0.279}}&0.324	& 0.283&0.325	& 0.288&0.329	& 0.280	&\textbf{{0.322}}\\
\midrule

\multirow{4}{*}{ETTh1} 
& 96  &	0.371	&\textbf{{0.397}}& 0.373& 0.398	& 0.371	&0.399	&0.386	&0.414	&\textbf{{0.370}}	&\textbf{{0.397}}\\
& 192 & \textbf{{0.411}}	&\textbf{{0.421}}	&0.425	&0.423	&\textbf{{0.411}}	&\textbf{{0.422}}	&0.434	&0.438	& 0.417&0.422\\
& 336 & \textbf{{0.448}}	&\textbf{{0.441}}	&0.469	&0.443	&0.454	&0.444	&0.467	&0.456	&0.454	&\textbf{{0.441}}\\
& 720 & \textbf{{0.441}}	&\textbf{{0.460}}	&0.465	&0.467	&0.447	&0.463	&0.447	&0.464	&0.461	&0.468\\
& Avg &  \textbf{{0.418}}	&\textbf{{0.430}}& 0.433&0.433	& 0.421&0.432 & 0.433&0.443  &0.425  &0.432\\
\midrule

\multirow{4}{*}{ETTh2} 
& 96  & \textbf{{0.278}}&	\textbf{{0.338}}	& 0.296&0.348	&0.287	&0.342	&0.316	&0.363	&0.289	&0.343\\
& 192 & \textbf{{0.351}}&	\textbf{{0.389}}&0.369	&0.399	&0.367	&0.397	&0.395	&0.412	&0.371	&0.404\\
& 336 & \textbf{{0.358}}&	\textbf{{0.398}}	&0.385	&0.414	&0.362	&0.400	&0.404	&0.427	&0.372	&0.407\\
& 720 & 0.404&	0.433	&0.406	&0.432	&\textbf{{0.402}}	&\textbf{{0.431}}	&0.422	&0.443	&0.418	&0.437\\
& Avg &	\textbf{{0.348}}&	\textbf{{0.390}}& 0.364&0.398	& 0.355&0.392 & 0.384	&0.411	&0.363	&0.398\\
\midrule

\multirow{4}{*}{Weather} 
& 96  &  0.162&	0.210 & 0.169&0.213	& 0.167&0.212	&0.167	&0.214	&{0.161}	&{0.208}\\
& 192 &  \textbf{{0.211}}&	\textbf{{0.253}}	&0.214	&0.259	&0.212	&{0.253}	&0.212	&0.255	&{0.211}	&0.254\\
& 336 &  \textbf{{0.267}}	&\textbf{{0.293}}	&0.277	&0.299	&0.276	&0.298	&0.270	&0.296	&0.278	&0.301\\
& 720 & \textbf{{0.335}}&	0.346 	&0.340	&0.349	&0.344	&{0.345}	&0.348	&0.355	&0.347	&\textbf{{0.345}}\\
& Avg &  \textbf{{0.244}}&	\textbf{{0.275}}	& 0.250&0.280	& 0.249	& 0.277	&0.249	&0.280	&0.249	& 0.277\\
\midrule

\multirow{4}{*}{Exchange Rate} 
& 96  &  	\textbf{{0.085}}&	0.205& \textbf{{0.085}}&\textbf{{0.204}}	& \textbf{{0.085}}	&0.205	&0.099	&0.224	& 0.086	&0.207\\
& 192 & 0.172&	0.296	&\textbf{{0.171}}	&\textbf{{0.294}}	&0.174	&0.298	&0.182	&0.307	&0.173	&0.298\\
& 336 & \textbf{{0.318}}&	\textbf{{0.408}}	&0.327	&0.416	&0.321	&0.412	&0.347	&0.433	&0.335	&0.421\\
& 720 & \textbf{{0.836}}&	\textbf{{0.696}}	&0.914	&0.726	&0.933	&0.737	&0.955	&0.744	&0.898	&0.723\\
& Avg & \textbf{{0.353}}&	\textbf{{0.401}}& 0.374&0.410	& 0.378&0.413	& 0.396&0.427	& 0.373&0.412\\
\midrule

\multirow{4}{*}{ILI} 
& 24  & \textbf{{1.583}}&\textbf{{0.802}} & 1.618&0.813	& 1.676&0.807	&2.764	&1.089	& 1.603	&0.793\\
& 36 & 1.601&	0.820	&1.683	&0.836	&\textbf{{1.510}}	&\textbf{{0.806}}	&1.967	&0.911	&1.643	&0.823\\
& 48 & 1.718&	0.815	& 1.865	&0.948	&2.114	&0.935	&1.955	&0.940	&\textbf{{1.638}}	& \textbf{{0.812}}\\
& 60 & \textbf{{1.920}}&	\textbf{{0.901}}	&1.977	&0.939	&1.955	&0.912	&2.018	&0.938	&2.013	&0.919\\
& Avg & \textbf{{1.705}}	&\textbf{{0.835}} & 1.786&0.884	& 1.813&0.865	&2.176	&0.969	&1.724	&0.837\\
\midrule

\multicolumn{2}{c|}{\textbf{1st Count}} &   \textbf{{23}}&\textbf{{22}}  & 5&5 & 5&5 &  0&0 & 6&9     \\

\bottomrule
\end{tabular}
\caption{Full ablation analysis of T3Time across seven datasets and forecast horizons. We evaluate the impact of removing four core components—Frequency Encoding, Multi-Head Cross-Modal Alignment, Channel-Wise Residual Connection, and Horizon-Aware Gating. Metrics reported are MSE and MAE (lower is better). \textbf{Bold}: the best.}
\label{tab:design_choice_full}
\end{table*}

\section{Appendix E: Design Variant}

Table \ref{tab:design_choice_full} presents a comprehensive ablation study across benchmark datasets and forecasting horizons, evaluating the contributions of key components in our T3Time architecture. We systematically remove four major design modules—Frequency Encoding, Multi-Head Cross-Modal Alignment, Channel-Wise Residual Connection, and Horizon-Aware Gating—to assess their individual impact on model performance.

Our full model demonstrates the most robust predictive capability, achieving the best performance in 23 out of 35 MSE cases and 21 out of 35 MAE cases, significantly outperforming all ablated variants. Among the variants, the model without the Gating Mechanism exhibits relatively competitive results, attaining the best MSE in 6 cases and best MAE in 9 cases. However, removing other modules—particularly the Residual Connection—leads to consistent and substantial performance degradation, indicating their critical role in maintaining deep temporal and semantic alignment. Averaged over all datasets and prediction lengths, the complete T3Time model achieves the lowest MSE and MAE, underscoring the contribution of each component and validating our architectural design choices.


\begin{figure*}[!t]
    \centering
    \includegraphics[width=0.9\textwidth]{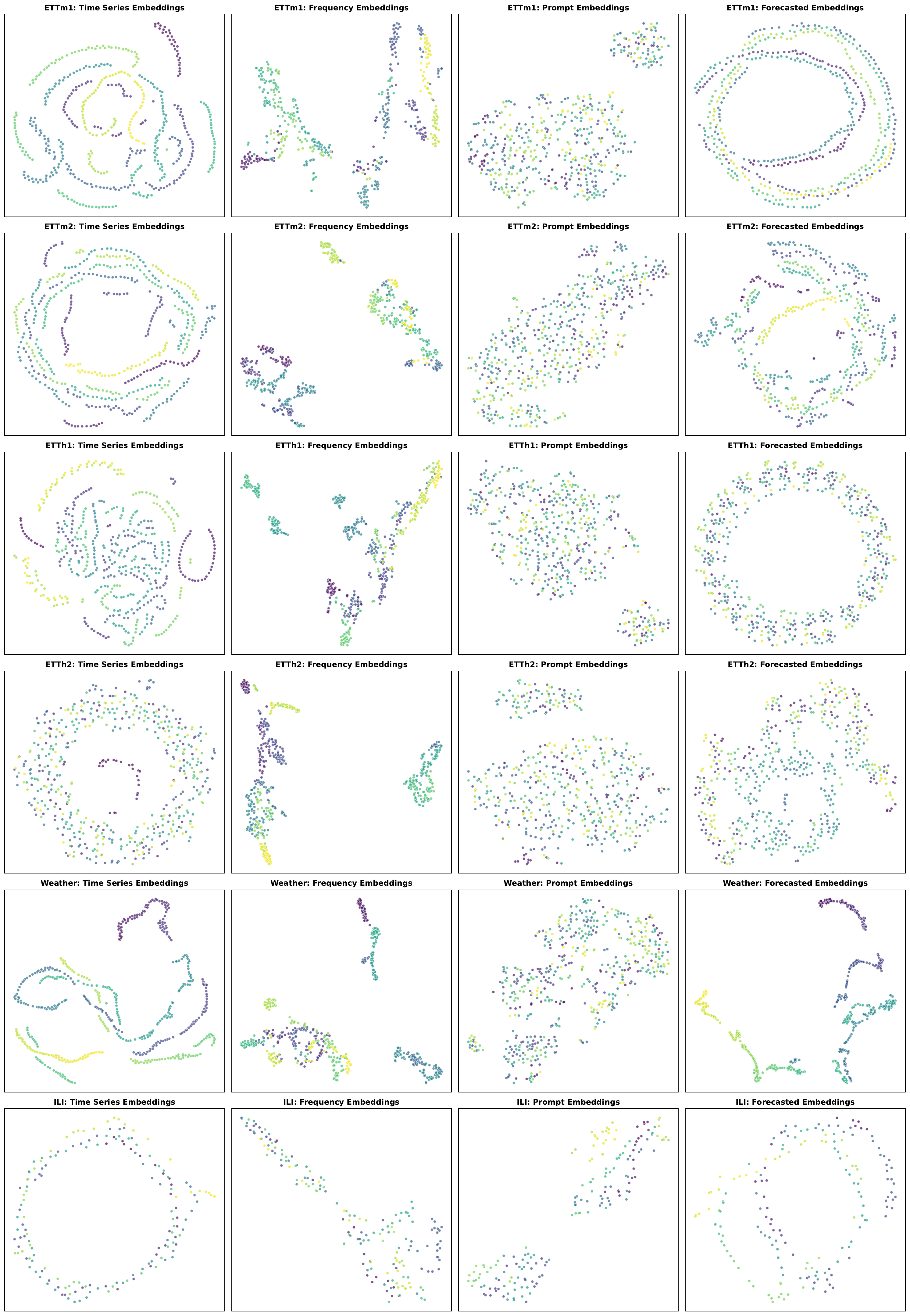}
    \caption{t-SNE visualization of four types of learned embeddings (time series, frequency, prompt, and forecasted) across six datasets. The prompt and forecasted embeddings exhibit clear, coherent clustering, while the time series and frequency embeddings appear more fragmented.}
    \label{fig:t-sne seperate}
\end{figure*}

\section{Appendix F: t-SNE Visualization}
\label{appendix:t-SNE}

Figure~\ref{fig:t-sne seperate} illustrates a comprehensive t-SNE visualization of the learned embeddings across six benchmark datasets (ETTm1, ETTm2, ETTh1, ETTh2, Weather, and ILI). For each dataset, we visualize four key embedding types extracted at different stages of the model: time series embeddings, frequency-domain embeddings, prompt embeddings, and forecasted (output) embeddings.

The time and frequency embeddings generally exhibit more dispersed or fragmented clustering, reflecting the challenge of modeling complex temporal and periodic patterns in isolation. In contrast, the prompt embeddings form denser, well-separated clusters, indicating that the LLM-encoded prompts inject strong semantic structure into the latent space. Notably, the forecasted embeddings display smooth, often spiral or circular manifolds, suggesting that the complete model learns compact and coherent representations that effectively align multiple modalities through cross-modal attention and residual refinement.

\end{document}